\pdfoutput=1

\documentclass[11pt]{article}

\usepackage[final]{acl}
\usepackage{comment}
\usepackage{times}
\usepackage{latexsym}
\usepackage{soul}

\usepackage[most]{tcolorbox}
\newtcbox{\myhighlight}{on line, colback=yellow!30, colframe=yellow!80!black, boxrule=0.5pt}

\usepackage[T1]{fontenc}

\usepackage[utf8]{inputenc}
\usepackage{amsmath, amsthm, amssymb}
\usepackage{amsthm}
\usepackage[shortlabels]{enumitem}
\usepackage{todonotes}

\usepackage{microtype}

\usepackage{inconsolata}

\usepackage{graphicx, caption, subcaption}
\usepackage{rotating}

\usepackage{booktabs}
\usepackage{multirow}

\title{XAutoLM: Efficient Fine-Tuning of Language Models via Meta-Learning and AutoML}

\author{
  \textbf{Ernesto L. Estevanell-Valladares\textsuperscript{1,2}},
  \textbf{Suilan Estévez-Velarde\textsuperscript{2}},
  \\
  \textbf{Yoan Gutiérrez\textsuperscript{1}},
  \textbf{Andrés Montoyo\textsuperscript{1}},
  \textbf{Ruslan Mitkov\textsuperscript{3}},
\\
  \textsuperscript{1}University of Alicante,
  \textsuperscript{2}University of Havana,
  \textsuperscript{3}University of Lancaster,
\\
  \small{
    \href{mailto:ernesto.estevanell@ua.es}{ernesto.estevanell@ua.es}
  }
}

\begin{document}
\maketitle

\begin{abstract}

Experts in machine learning leverage domain knowledge to navigate decisions in model selection, hyperparameter optimization, and resource allocation. This is particularly critical for fine-tuning language models (LMs), where repeated trials incur substantial computational overhead and environmental impact. However, no existing automated framework simultaneously tackles the entire model selection and hyperparameter optimization (HPO) task for resource-efficient LM fine-tuning. We introduce \textbf{XAutoLM}, a meta-learning-augmented AutoML framework that reuses past experiences to optimize discriminative and generative LM fine-tuning pipelines efficiently. XAutoLM learns from stored successes and failures by extracting task- and system-level meta-features to bias its sampling toward valuable configurations and away from costly dead ends.  On four text classification and two question-answering benchmarks, XAutoLM surpasses zero-shot optimizer's peak $F1$ on five of six tasks, cuts mean evaluation time of pipelines by up to 4.5x, reduces search error ratios by up to sevenfold, and uncovers up to 50\% more pipelines above the zero-shot Pareto front. In contrast, simpler memory-based baselines suffer negative transfer. We release XAutoLM and our experience store to catalyze resource-efficient, Green AI fine-tuning in the NLP community.

\end{abstract}

\section{Introduction}

Fine-tuning large language models (LLMs) has become indispensable across natural language processing (NLP) applications, yet even “small” models such as BERT \citep{devlin2018bert} or T5 \citep{t5} incur substantial computational cost and carbon emissions \citep{wang2023energy,schwartz2020green}. Rather than exhaustively evaluating every model and hyperparameter combination, human experts draw on domain knowledge to focus on promising regions of this vast design space.

Automated Machine Learning (AutoML) seeks to mimic expert intuition by automating the two core stages of pipeline construction, model selection (MS) and hyperparameter optimization (HPO), into a unified search loop \citep{hutter2019automated}. AutoML techniques have matured in areas such as tabular and vision tasks \cite{hutter2019automated}, showing competitive performance against human experts \cite{estevez2020automatic}. However, the joint MS+HPO pipeline for language models presents an ample, mixed discrete-continuous search space whose repeated evaluations are prohibitively costly \citep{wang2023energy}, thus posing a significant challenge for automation. While several recent efforts address HPO for LMs in isolation \citep{priorband}, surveys highlight the underdevelopment of full-pipeline AutoML in NLP \citep{automl-llm-survey-opinion}, and no framework systematically unifies model selection and HPO under tight compute and Green AI constraints.

To address these shortcomings, we present \textbf{XAutoLM}, an AutoML framework that unifies model selection and hyperparameter optimization for LM fine-tuning via meta-learning. XAutoLM constructs an \emph{experience-aware prior} from a repository of past pipeline evaluations annotated with task- and system-level meta-features which steers the search toward historically promising and away from infeasible configurations. Empirically, across four classification and two question-answering benchmarks, our method yields pipelines with stronger performance-time trade-offs than zero-shot or naive baselines under identical wall-clock budgets (Tables~\ref{tab:multi-objective},~\ref{tab:qa}). We release the code and the full experience store\footnote{\url{https://github.com/EEstevanell/XAutoLM}} to support sustainable, reproducible LM fine-tuning in the NLP community.

We summarize our main contributions as follows:
\begin{itemize}
  \item A unified, meta-learning–augmented AutoML framework that integrates \emph{both} model selection and hyperparameter optimisation for discriminative and generative LM fine-tuning.
  
  \item An extensible, task- and model-agnostic \emph{experience-aware prior} that conditions the search on task \emph{and} system meta-features and explicitly leverages negative traces to avoid costly dead ends.
  
  \item A comprehensive evaluation on six benchmarks showing consistent gains in $F_1$, mean pipeline evaluation time, and error ratio, and stronger Pareto fronts than zero-shot and naive memory baselines (see Section~\ref{sec:experiments}; Tables~\ref{tab:multi-objective},~\ref{tab:qa}).
\end{itemize}

We next review related work (Section~\ref{sec:sota}), present XAutoLM (Section~\ref{sec:proposal}), and report the experimental setup and results (Section~\ref{sec:experiments}), followed by analysis (Section~\ref{sec:discussion}) and, finally, conclusions and limitations (Sections~\ref{sec:conclusions},~\ref{sec:limitations}).

\section{Related Work} \label{sec:sota}

AutoML strategies in language modelling can be divided into two (not necessarily disjoint) subsets: AutoML for LLMs and LLMs for AutoML~\citep{automl-llm-survey-opinion}. The former comprises AutoML techniques to produce optimal LM pipelines tailored for specific scenarios, akin to traditional AutoML. The latter employs language models to enhance the AutoML process, for example, by providing linguistic interfaces to configure the optimisation process or leveraging them to guide the search (e.g., using LMs to generate code for optimal ML pipelines).

\textbf{AutoML for LLMs} in particular poses significant challenges \cite{automl-llm-survey-opinion}. Namely, LMs are extremely resource-intensive \cite{bannour2021evaluating}, even when only considering their later stages (e.g., fine-tuning, inference)\label{sec:sota:challenge-2}. Table~\ref{tab:sota-features} compares AutoML approaches that leverage LLMs according to relevant features characterising their responses to the field's challenges. 

\begin{table}[ht!]
\centering{
\resizebox{.9\linewidth}{!}{
    \begin{tabular}{rcccccccc}
        \multicolumn{1}{r|}{Systems\ \ \begin{sideways}\ \ \ \ Features\end{sideways}}  & \multicolumn{1}{r|}{\begin{sideways}AutoML for LLMs\end{sideways}} &
        \multicolumn{1}{r|}{\begin{sideways}LLMs for AutoML\end{sideways}} & \multicolumn{1}{r|}{\begin{sideways}Inference\end{sideways}}    & \multicolumn{1}{r|}{\begin{sideways}Fine-tuning\end{sideways}}  & \multicolumn{1}{r|}{\begin{sideways}HPO\end{sideways}}   & \multicolumn{1}{r|}{\begin{sideways}Model Selection\end{sideways}} & \multicolumn{1}{r}{\begin{sideways}Meta-learning\end{sideways}}\\ 

        \hline
        \multicolumn{1}{r|}{APE}       & \multicolumn{1}{c|}{}  & \multicolumn{1}{c|}{$\checkmark$}  & \multicolumn{1}{c|}{$\checkmark$} & \multicolumn{1}{c|}{} & \multicolumn{1}{c|}{} & \multicolumn{1}{c|}{}  & \multicolumn{1}{c}{}\\ 
        
        \hline
        \multicolumn{1}{r|}{GPT-NAS}   & \multicolumn{1}{c|}{$\checkmark$} & \multicolumn{1}{c|}{$\checkmark$}   & \multicolumn{1}{c|}{} & \multicolumn{1}{c|}{} & \multicolumn{1}{c|}{$\checkmark$} & \multicolumn{1}{c|}{$\checkmark$}  & \multicolumn{1}{c}{}\\ 
        
        \hline
        \multicolumn{1}{r|}{GL-Agent} & \multicolumn{1}{c|}{} & \multicolumn{1}{c|}{$\checkmark$}    & \multicolumn{1}{c|}{} &
        \multicolumn{1}{c|}{} & \multicolumn{1}{c|}{} & \multicolumn{1}{c|}{} & \multicolumn{1}{c}{}\\  
        
        \hline
        \multicolumn{1}{r|}{AutoGen}   & \multicolumn{1}{c|}{$\checkmark$}  & \multicolumn{1}{c|}{$\checkmark$}  & \multicolumn{1}{c|}{$\checkmark$} & \multicolumn{1}{c|}{} & \multicolumn{1}{c|}{} & \multicolumn{1}{c|}{}   & \multicolumn{1}{c}{}\\ 
        
        \hline
        \multicolumn{1}{r|}{EcoOptiGen}     & \multicolumn{1}{c|}{$\checkmark$} & \multicolumn{1}{c|}{}   & \multicolumn{1}{c|}{$\checkmark$} & \multicolumn{1}{c|}{} & \multicolumn{1}{c|}{$\checkmark$} & \multicolumn{1}{c|}{}   & \multicolumn{1}{c}{}\\ 
        
        \hline
        \multicolumn{1}{r|}{AutoML-GPT}     & \multicolumn{1}{c|}{$\checkmark$} & \multicolumn{1}{c|}{$\checkmark$}   & \multicolumn{1}{c|}{} & \multicolumn{1}{c|}{} & \multicolumn{1}{c|}{$\approx$}    & \multicolumn{1}{c|}{} & \multicolumn{1}{c}{}\\ 
        
        \hline
        \multicolumn{1}{r|}{HuggingGPT}     & \multicolumn{1}{c|}{$\approx$} & \multicolumn{1}{c|}{$\checkmark$}      & \multicolumn{1}{c|}{$\checkmark$} & \multicolumn{1}{c|}{} & \multicolumn{1}{c|}{} & \multicolumn{1}{c|}{$\checkmark$} & \multicolumn{1}{c}{}\\ 
        
        \hline
        \multicolumn{1}{r|}{AutoM3L}   & \multicolumn{1}{c|}{} & \multicolumn{1}{c|}{$\checkmark$}   & \multicolumn{1}{c|}{} & \multicolumn{1}{c|}{} & \multicolumn{1}{c|}{$\checkmark$} & \multicolumn{1}{c|}{$\checkmark$} & \multicolumn{1}{c}{$\approx$}\\ 
        
        \hline
        \multicolumn{1}{r|}{PriorBand} & \multicolumn{1}{c|}{$\checkmark$} & \multicolumn{1}{c|}{}   & \multicolumn{1}{c|}{} & \multicolumn{1}{c|}{$\checkmark$} & \multicolumn{1}{c|}{$\checkmark$} & \multicolumn{1}{c|}{} & \multicolumn{1}{c}{$\checkmark$}\\ 
        
        \hline
        \multicolumn{1}{r|}{GizaML}    & \multicolumn{1}{c|}{}  & \multicolumn{1}{c|}{$\checkmark$}  & \multicolumn{1}{c|}{} & \multicolumn{1}{c|}{} & \multicolumn{1}{c|}{$\checkmark$} & \multicolumn{1}{c|}{$\checkmark$}    & \multicolumn{1}{c}{$\checkmark$}\\ 

        \hline
        \multicolumn{1}{r|}{GE} & \multicolumn{1}{c|}{$\checkmark$}  & \multicolumn{1}{c|}{$\checkmark$}  & \multicolumn{1}{c|}{$\checkmark$} & \multicolumn{1}{c|}{} & \multicolumn{1}{c|}{$\checkmark$} & \multicolumn{1}{c|}{} & \multicolumn{1}{c}{$\approx$}\\ 
        
        \hline
        \multicolumn{1}{r|}{AutoGOAL} & \multicolumn{1}{c|}{$\checkmark$}  & \multicolumn{1}{c|}{}  & \multicolumn{1}{c|}{$\checkmark$} & \multicolumn{1}{c|}{} & \multicolumn{1}{c|}{$\checkmark$} & \multicolumn{1}{c|}{$\checkmark$}     & \multicolumn{1}{c}{}\\ 
        
        \hline
        \multicolumn{8}{l}{\textit{Introduced in this paper}}\\ 
        
        \hline
        \multicolumn{1}{r|}{\textbf{XAutoLM}} & \multicolumn{1}{c|}{$\checkmark$} & \multicolumn{1}{c|}{}   & \multicolumn{1}{c|}{$\checkmark$} & \multicolumn{1}{c|}{$\checkmark$} & \multicolumn{1}{c|}{$\checkmark$} & \multicolumn{1}{c|}{$\checkmark$}     & \multicolumn{1}{c}{$\checkmark$}\\
    \end{tabular}
}}
\caption{Comparison of systems for AutoML with LLMs}
\label{tab:sota-features}
\end{table}

We observe that there are more \textbf{LLMs for AutoML} systems than vice versa, likely due to the proliferation of prompt engineering and increased access to open-source LMs. For instance, \citet{ape} developed the Automatic Prompt Engineer (APE) system, which achieved performance competitive with human-generated instructions. In contrast, systems such as GL-Agent \cite{autograph}, AutoM3L \cite{autom3l} and GizaML \cite{gizaml} integrate language models into their optimization strategies to produce graph learning pipelines, highly capable multi-modal ML pipelines, and time-series forecasting pipelines, respectively.

Systems like AutoGen \cite{autogen}, GPT-NAS \cite{gptnas}, GE \cite{morris2024llm}, AutoML-GPT \cite{automl-gpt}, and HuggingGPT \cite{hugginggpt} are hybrids that span both categories; they leverage LMs to produce LM-based solutions. However, the last two differ from traditional AutoML (and NAS) systems: AutoML-GPT does not evaluate solution candidates (only simulates their training), and HuggingGPT produces responses to prompts without outputting the pipelines capable of handling them.

Often, the choice of model is as, if not more, critical than the hyperparameter configuration used to produce responses. We found that AutoGOAL \cite{estevanell2024balancing} optimizes pipelines by balancing efficiency and performance metrics, taking into account both model selection and HPO, but only supports LMs for inference. All other AutoML for LLMs systems we surveyed, such as EcoOptiGen \cite{ecooptigen} and PriorBand \cite{priorband}, focus solely on HPO. 

Nonetheless, we find no single framework that simultaneously addresses model selection and hyperparameter optimization for LM fine-tuning, particularly when resource limitations exist. 

\section{Proposal} \label{sec:proposal}

We introduce \textbf{XAutoLM}, the first AutoML framework that unifies \emph{model selection} and \emph{hyperparameter optimisation} for both \textbf{discriminative} and \textbf{generative} language model fine-tuning. Our pipelines are composed of (i) a base LM from a curated pool of encoders and generators (Table~\ref{tab:llms}), (ii) one of three fine-tuning strategies; full, partial, or LoRA~\citep{lora}, and (iii) a hyperparameter configuration. XAutoLM jointly explores this mixed search space by reusing past \textit{experiences}, e.g., ``LoRA-tuned DistilBERT achieved high macro-F1 on SST-2 under low VRAM'', to steer the optimizer toward high-utility regions and away from error-prone configurations. This holistic reuse enables XAutoLM to discover strong fine-tuning pipelines under tight compute budgets.

\begin{table}[ht!]
        \centering
        \resizebox{0.95\columnwidth}{!}{
        \begin{tabular}{@{}l@{}}
\toprule
\textbf{\textit{Discriminative}}    \\ \midrule
BERT \cite{devlin2018bert}           \\
DistilBERT \cite{sanh2020distilbert} \\
RoBERTa \cite{roberta}               \\
XLM-RoBERTa \cite{xlm-roberta}       \\
DeBERTa \cite{deberta}               \\
DeBERTaV3 \cite{debertav3}           \\
MDeBERTaV3 \cite{debertav3}          \\
ALBERT-v1 \cite{albertv1}            \\
ELECTRA \cite{electra}               \\ \midrule
\textbf{\textit{Generative}}        \\ \midrule
T5 \cite{t5}                         \\
FLAN-T5 \cite{flan-t5}               \\
GPT-2 \cite{gpt2}                    \\
PHI-3 \cite{abdin2024phi3technicalreporthighly} \\ \midrule
\textbf{\textit{New Additions}}        \\ \midrule
PHI-3.5 (Mini-Inst) \cite{abdin2024phi}               \\
PHI-4 (Mini-Inst, Reasoning) \cite{abdin2024phi}                   \\
MIXTRAL (8x7B) \cite{mixtral-experts}                   \\
MISTRAL NEMO (Base-Inst) \cite{mistral-nemo}                  \\
Llama 3.1, 3.2 (1B - 70B)  \cite{grattafiori2024llama3herdmodels}         \\
DeepSeek R1 \cite{deepseekai2025deepseekr1incentivizingreasoningcapability}  \\ \bottomrule
\end{tabular}}
\caption{LMs available in AutoGOAL's algorithm pool.}
\label{tab:llms}
\end{table}

\paragraph{Background}
XAutoLM builds on AutoGOAL’s\footnote{Open-source available at: \url{https://github.com/autogoal/autogoal}, licensed without restriction.} probabilistic optimizer \citep{estevez2020automatic}. The optimizer represents every valid LM pipeline \(c\) as a point in a \emph{mixed} search space that combines discrete choices (e.g.\ fine-tuning method, model, tokenizer) with continuous hyperparameters (e.g.\ learning rate, dropout). It maintains a probability distribution \(P(c\!\mid\!\theta)\) over that space. It repeats a simple \textit{sample–evaluate–update} loop: (1) \emph{sample} a batch of pipelines from \(P(c\!\mid\!\theta)\); (2) \emph{evaluate} them on the target task; and (3) \emph{update} \(P(c\!\mid\!\theta)\) so that high-performing pipelines gain probability mass while under-performing and failures lose it. AutoGOAL always \emph{initializes} this distribution \textbf{uniformly}, meaning every pipeline, adequate or not, is equally likely at the first generation.

\subsection{Process Overview}

XAutoLM replaces this uniform cold start with an \emph{experience-aware prior} that follows a structured meta-learning process. Initially, the framework retrieves relevant historical evaluations (experiences) from a centralized repository (Section ~\ref{sec:exp-store}). Then, it computes detailed task and system meta-features (Section ~\ref{sec:meta-features}) to characterize the complexity and available resources for the present optimisation task. Leveraging this information, XAutoLM probabilistically adjusts the AutoML search space (Section ~\ref{sec:warmstart}), focusing on historically successful configurations and reducing exploration of previously unsuccessful paths. Once configured, the AutoML optimisation starts, fine-tuning pipelines are evaluated, and their outcomes, both successful and unsuccessful, are recorded back into the experience repository, to be used in future runs.

\subsection{Experience Store} \label{sec:exp-store}

Our system learns from a growing repository of \textit{experiences}; past pipeline evaluations that capture every factor influencing performance. Formally, an experience is a 4-tuple \(e=\langle c,\,\mathbf{m},\,t,\,s\rangle\) where
\(c\) is the complete pipeline configuration, \(\mathbf{m}\) the vector of recorded metrics (e.g.\ $\mathrm{F1}$, ROUGE, evaluation time), \(t\) a task meta-feature vector, and \(s\) straightforward system descriptors such as CPU cores, RAM, and GPU memory. 

We label an experience \textbf{positive} if all fitness metrics are valid and \textbf{negative} otherwise, usually due to errors occurring during evaluation (out-of-memory, timeout, etc.). Both types are essential: positives pull the search toward valuable regions, and negatives push it away from costly dead-ends (Section ~\ref{sec:warmstart}).

\subsubsection{Meta-Features} \label{sec:meta-features}

We design two complementary meta-feature templates according to the \emph{nature of the output space} of a task. When the output is drawn from a \textbf{closed label set}, as in text classification or sequence labelling, dataset difficulty is dominated by class imbalance and document-length variation. Conversely, tasks whose output is an \textbf{open text sequence} (question answering, summarisation, translation) demand features that capture the relationship between the input prompt and the target text. Table~\ref{tab:task-features} lists the core features for each template; the same templates can be reused for other label-based or free-form generation tasks with minimal adaptation.

\begin{table}[ht]
\centering
\subfloat[\textbf{Label-based}]{
\resizebox{.39\linewidth}{!}{
\begin{tabular}{l}
\hline
\textbf{\textit{Category}}/Feature                   \\ \hline
\multicolumn{1}{c}{\textit{\textbf{Dataset}}}        \\
Nr Samples                                           \\
Nr Classes                                           \\
Entropy                                              \\
Min Cls Prob                                         \\
Max Cls Prob                                         \\
Imbalance Ratio                                      \\ \hline
\multicolumn{1}{c}{\textit{\textbf{Documents}}}      \\
Avg. Length                                          \\
Std. Length                                          \\
Coef. Var. Length                                    \\ \hline
\multicolumn{1}{c}{\textit{\textbf{Landmark}}}       \\
PCA + D.Tree Acc.                                    \\ \hline
\end{tabular}
}}\hfill
\subfloat[\textbf{Generation}]{
\resizebox{.48\linewidth}{!}{
\begin{tabular}{l}
\hline
\multicolumn{1}{c}{\textbf{\textit{Category}}/Feature} \\ \hline
\multicolumn{1}{c}{\textit{\textbf{Dataset}}}       \\
Nr Samples                                          \\ \hline
\multicolumn{1}{c}{\textit{\textbf{Prompt}}}        \\
Avg.\textbackslash Len (chars)                      \\
Std.\textbackslash Len                              \\
Lexical Diversity (TTR)                             \\\hline
\multicolumn{1}{c}{\textit{\textbf{Prompt–Target}}} \\
Avg.\textbackslash Len Ratio (T/P)                  \\
Vocabulary Novelty                                  \\
Semantic Similarity                                 \\
ROUGE-L F1                                          \\ \hline
\multicolumn{1}{c}{\textit{\textbf{Semantic}}}      \\
Mean Prompt Embedding                               \\ \hline
\end{tabular}}}
\caption{Representative task meta-features.}
\label{tab:task-features}
\end{table}

Experiences record a minimal hardware profile in \emph{s} (CPU cores, CPU frequency, total RAM, GPU VRAM) so similarity and feasibility reflect both task and system characteristics. For instance, while Llama 3.1 70B may yield superior results to smaller alternatives, systems with low VRAM cannot utilize its power.

XAutoLM constructs a holistic representation of each optimization scenario by combining task-specific and system-level meta-features, enabling robust similarity assessments across diverse contexts.

\subsection{Warm-Start optimization} \label{sec:warmstart}

XAutoLM maintains a probabilistic model \(P(c \mid \theta)\) \cite{estevez2020automatic} over pipeline configurations \(c\). When a new task~\(T\) arrives, we retrieve a set of past \emph{experiences} \(\mathcal{E}=\{e_1,\dots,e_n\}\) and update the model in two sweeps; one for positive experiences, one for negatives:

\begingroup
\small
\setlength{\jot}{1pt}
\begin{align}
P(c \mid \theta) &\leftarrow (1-\alpha_i^{+}) \, P(c \mid \theta) + \alpha_i^{+} \, P_i(c \mid \theta),\label{eq:prob-update-pos}\\[1ex]
P(c \mid \theta) &\leftarrow (1+\alpha_i^{-}) \, P(c \mid \theta) - \alpha_i^{-} \, P_i(c \mid \theta)\label{eq:prob-update-neg}
\end{align}
\endgroup


where \(P_i(c\mid\theta)\) is the empirical distribution induced by configuration \(c\) in experience \(e_i\). Therefore \emph{pull} the search toward successful regions and \emph{push} it away from unsuccessful ones.  The strength of each pull/push is governed by the
\emph{learning rates} \(\alpha_i^{+}\) and \(\alpha_i^{-}\).

We compute experience-specific learning rates considering their similarity to the current task and historical performance. Specifically, these rates are computed as follows:

\begin{align}
\alpha_i^{+} &= \alpha_{\max}^{+}\;u_i\;e^{-\beta\,d_i}, \label{eq:alpha-pos}\\[1ex]
\alpha_i^{-} &= \alpha_{\max}^{-}\;e^{-\beta\,d_i}. \label{eq:alpha-neg}
\end{align}

Here \(\alpha_{\max}^{+}\) and \(\alpha_{\max}^{-}\) are predefined maximum learning rates, \(u_i \in [0,1]\) is a utility score (defined below) assigned \emph{only} to positive experiences, and \(d_i\) is the distance between the current task and the one that generated experience~\(e_i\). The exponential kernel \(e^{-\beta d_i}\) down-weights experiences that are less similar to the current task; \(\beta>0\) is an adaptive decay factor.

\paragraph{Task Similarity.}
Each task is described by a meta-feature vector \(t\). Similarity is measured with a distance \(d_i = \mathrm{Dist}(t_{T},t_i)\) (e.g., Euclidean or Cosine). \(\beta\) is set automatically to compensate for scale:

\begin{align}
\beta = \frac{\beta_{\text{scale}}}{\sigma_d + \varepsilon},
\quad
\sigma_d = \mathrm{Std}\bigl(\{d_1,\dots,d_n\}\bigr),
\end{align}

where \(\varepsilon \!\!>\! 0\) prevents division by zero.

\paragraph{Utility Score.} 
The utility function \(u_i\) quantifies the quality of each positive experience \(e_i\) relative to others from the same task. XAutoLM supports three distinct utility computation strategies: (\textit{i})~Weighted Sum, (\textit{ii})~Linear Front, and (\textit{iii})~Logarithmic Front:

\subparagraph{Weighted Sum.} 
Let \(\mathcal{M}\) denote the set of recorded performance metrics for each experience, such as F1, accuracy, evaluation time, or ROUGE-L. Each metric \(m \in \mathcal{M}\) is associated with a known optimisation direction (maximize or minimize) and an importance weight \(w_m\). For each positive experience \(e_i\), we first normalize its metric value \(m_i\):

\begingroup
\small
\setlength{\jot}{1pt}
\begin{align}
m_i' = 
\begin{cases}
\dfrac{m_i - m_{\min}}{m_{\max} - m_{\min}}, & \text{if maximized},\\[1.4ex]
1 - \dfrac{m_i - m_{\min}}{m_{\max} - m_{\min}}, & \text{if minimized},
\end{cases}
\end{align}
\endgroup

where \(m_{\min}\) and \(m_{\max}\) denote the minimum and maximum values observed across all positive experiences for the metric \(m\). If all metric values are identical, we default to a neutral utility score of 0.5 to avoid division by zero. The overall weighted utility score is computed as:

\begin{align}
u_i = \frac{\sum_{m \in \mathcal{M}} w_m \cdot m_i'}{\sum_{m \in \mathcal{M}} w_m},
\end{align}

\subparagraph{Linear Front.} 
In the Linear Front utility scheme, we first apply non-dominated sorting (NSGA-II style~\citep{nsgaii}) to all positive experiences, creating \(N\) Pareto fronts based on the recorded metrics in \(\mathcal{M}\). Experiences in front 0 are non-dominated, followed by those in front 1, and so forth. Each positive experience \(e_i\) in front \(f_i\) is assigned a utility score inversely proportional to its front rank:

\begin{align}
u_i = \frac{N - f_i}{N},
\end{align}

\subparagraph{Logarithmic Front.} 
Using non-dominated sorting, the Logarithmic Front approach similarly ranks experiences into \(N\) Pareto fronts. However, to amplify the distinction among the highest-performing experiences (i.e., those in lower-numbered fronts), utilities decrease logarithmically with rank:

\begin{align} 
u_i = \frac{\ln(N - f_i + 1)}{\ln(N + 1)}, \label{eq:log}
\end{align}

These three utility functions provide complementary strategies for prioritizing past experiences. This flexibility allows XAutoLM to adapt effectively across diverse AutoML scenarios.

\section{Experimentation} \label{sec:experiments}

We report results from two \emph{independent} transfer experiments designed to isolate knowledge reuse \emph{within} a task family.  
The first study targets \textbf{text classification}.  LIAR \cite{wang2017liar}, SST-2 \cite{sst2}, MELD \cite{poria2018meld} and AG~News \cite{agnews} present a deliberate gradient in sample size, label entropy, and average document length: LIAR (6 classes, 13k claims) and MELD (7 emotions, 14k utterances) are notoriously low-resource, whereas the polarity benchmark SST-2 (68k) and the large-scale news corpus AG (128k) approach the upper bound of single-GPU throughput. Previous work shows peak $F1_{\text{macro}}$ to vary from 0.23 (LIAR) to 0.93 (AG) \citep{REUSENS2024124302}, offering a realistic range for efficiency–performance trade-offs.  

The second experiment focuses on \textbf{question answering}.  We select SQuAD 1.1 \cite{rajpurkar2016squad} and DROP \cite{dua2019drop} because they share the same input modality yet differ sharply in answer type, extractive spans versus multi-step numerical reasoning, making them a challenging test-bed for generative pipelines.  For both studies, experiences are only exchanged among tasks of the same family; classification traces are invisible to QA runs and vice-versa.  This constraint ensures that the reported gains stem from \emph{task-relevant} meta-knowledge rather than accidental data leakage.

\paragraph{Hardware.}
All classification experiments run on an \texttt{i9-9900K} (16 threads, 35 GB RAM cap) paired with a single RTX TITAN (24 GB).
QA experiments require larger context windows and execute on an \texttt{AMD EPYC 7742} (64 threads, identical RAM cap) with an A100 40 GB.

\paragraph{Baselines.}
Every run is compared against \textbf{Zero-Shot AutoGOAL}, the original optimizer with a uniform sampling distribution; in this setting, the update rules of equations~(\ref{eq:prob-update-pos})–(\ref{eq:log}) are never triggered.

In the text classification study, we include a naive \textbf{kNN-50} memory baseline for comparing against a naive experience retrieval method. For every target task, we assemble a query vector that concatenates (a) the task meta-features, (b) the current system profile, and (c) the best metric values observed across all stored traces; this encourages the search to drift toward high-performing regions. Distances to \emph{}positive traces are computed on the full feature+metric space, whereas distances to \emph{negative} traces ignore metrics (errors lack valid scores). The $k$ nearest positives and $k$ nearest negatives are selected; all receive the same fixed learning rate $\alpha_i^{\pm}=1/k$. Setting $u_i\!=\!1$ and $\beta\!=\!0$ in equations~(\ref{eq:alpha-pos})–(\ref{eq:alpha-neg}) reduces our framework to this simple neighbour rule. For question answering the repository contains only between 5 and 10 positive traces per source task, making a neighbour count unreliable; therefore Zero-Shot remains the sole baseline in that study.

\paragraph{Warm-Start Priors.}

Throughout the paper, a \emph{pipeline configuration} is a concrete tuple \((\text{LM}, \text{fine-tuning recipe}, \text{hyperparameters})\) that the AutoML engine executes and evaluates. A \emph{warm-start prior} (WS prior) instead parameterizes the initial sampling distributions used by the meta-learner; it is defined by the distance type, utility scheme, decay factor \(\beta_{\text{scale}}\), and pull limits \((k_{\text{pos}}, k_{\text{neg}})\).


For each task, we enumerate \(\approx 180\) WS-prior parameterizations. For a given candidate prior to a task, we apply it with the fixed experience store (leaving the experience for the current task out) to obtain the \emph{induced} sampling distribution \(p\) over fine-tuning methods on that task. We then compute the total-variation (TV) distance between this induced marginal and the uniform distribution over the same method set. We rank candidates by TV and split them into three data-driven strata (\textit{low} | \textit{moderate} | \textit{high} bias) at prominent TV gaps (\(\approx2\times\)). In classification, we select per strata the median-TV and max-TV priors (six priors total). In QA, we select only the max-TV prior per strata (three priors) to respect the compute budget. Full probability plots of the induced method distributions and the selected prior identifiers are provided in Appendix~\ref{sec:appendix-prob-visualizations}.



\paragraph{Execution protocol.}

For each task, we first ran the Zero-shot configuration for 48 hours to populate the experience store. Table~\ref{tab:experiences} reports the positive/negative traces generated by this baseline run on each task. We then executed the kNN-50 baseline and all WS-prior variants for \textbf{24 hours of wall-clock time} each. The warm-start mechanism accesses only experiences originating from other tasks within the same study (clean cross-task transfer; see Table~\ref{tab:experiences}). For fairness in reporting, Zero-shot metrics are computed from the \textbf{first 24 hours} of their 48 hours runs, matching the wall-time allocated to WS-priors and kNN-50. This protocol isolates whether experience improves both effectiveness and efficiency under the same time budget.

\begin{table}[th!]
\centering 
\resizebox{0.9\columnwidth}{!}{
   \begin{tabular}{l|lll|lll}
\hline
\multirow{2}{*}{Dataset} & \multicolumn{3}{c|}{Generated} & \multicolumn{3}{c}{Available} \\ \cline{2-7} 
                         & Pos      & Neg     & Total     & Pos     & Neg     & Total     \\ \hline
LIAR                     & 100      & 236     & 336       & 116     & 480     & 596       \\
SST2                     & 33       & 122     & 155       & 183     & 594     & 777       \\
MELD                     & 68       & 190     & 258       & 148     & 526     & 674       \\
AG NEWS                  & 15       & 168     & 183       & 216     & 548     & 764       \\ \hline
SQUAD                    & 5        & 124     & 129       & 10      & 160     & 170       \\
DROP                     & 10       & 160     & 170       & 5       & 124     & 129       \\ \hline
\end{tabular}    
}
\caption{Disposition of experiences participating in the experiments.}
\label{tab:experiences}
\end{table}


In every AutoML run, each discovered LM pipeline has up to 1.5\,GPU-hours in Text Classification and 2\,GPU-hours in QA for evaluation. Objectives are $\langle F1_{\text{macro}}, ET \rangle$ for classification and $\langle F1, ET \rangle$ for QA, where $ET$ is the wall-clock evaluation time of a pipeline (in seconds). All searches share a fixed random seed ($42$) and the same hardware; therefore, differences arise solely from the chosen warm-start prior.

\subsection{Text Classification Results}

Table~\ref{tab:multi-objective} summarizes the effect of WS-priors on the four classification benchmarks. We report both performance and efficiency: max and mean $F1_{\text{macro}}$ reflect peak and average classification quality; mean evaluation time (ET) captures resource cost; the error ratio indicates the share of failed pipeline evaluations; and hypervolume (HV) measures Pareto-front coverage in objective space \citep{zitzler1998multiobjective}. Mean ET is averaged over successfully completed pipeline evaluations only (i.e., runs that return valid fitness metrics); failed evaluations (e.g., out-of-memory, timeouts, runtime errors) are excluded from ET and are accounted for by the error ratio. All methods are run under the \textbf{same 24 hours single-GPU budget} (cf.\ Execution protocol), so ET differences reflect pipeline runtime rather than total search compute.

\begin{table}[ht!]
\resizebox{1\linewidth}{!}{
\begin{tabular}{@{}p{0.30em}llllllll@{}}
\toprule
 & WS Prior & \begin{tabular}[c]{@{}l@{}}Max \\ $F1_m$\end{tabular} & \begin{tabular}[c]{@{}l@{}}Mean \\ $F1_m$\end{tabular} & \begin{tabular}[c]{@{}l@{}}Min \\ $ET$\end{tabular} & \begin{tabular}[c]{@{}l@{}}Mean \\$ET$ \end{tabular} & HV & \begin{tabular}[c]{@{}l@{}}No. \\ Eval\end{tabular} & \begin{tabular}[c]{@{}l@{}}Error \\ Ratio\end{tabular} \\ \midrule
\multirow{12}{*}{\rotatebox{90}{LIAR}} & Zero-shot & 0.24 & \textbf{0.10} & 12 & 537 & 0.06 & 202 & 0.73 \\
 & kNN (50) & 0.24  & \textbf{0.10}  & 28 & 451 & 0.11 & \textbf{240} & 0.44 \\
 & Low (LIAR) & \textbf{0.26} & \textbf{0.10} & 16 & 480 & 0.10 & 197 & 0.70 \\
 & Low (Med) & 0.25 & 0.09 & 31 & 380 & \textbf{0.36} & 220 & 0.69 \\
 & Low (Max) & 0.25 & 0.09 & 21 & 410 & 0.08 & 190 & 0.66  \\
 & Mod (LIAR) & \textbf{0.26} & \textbf{0.10} & 36 & 462 & 0.01 & 132 & 0.53 \\
 & Mod (Med) & 0.24 & \textbf{0.10} & 13 & 469 & 0.04 & 146 & 0.61 \\
 & Mod (Max) & 0.25 & 0.08 & 44 & 516 & 0.05 & 121 & 0.39  \\
 & High (LIAR) & 0.25 & \textbf{0.10} & \textbf{6} & \textbf{153} & 0.20& 302 & \textbf{0.09} \\
 & High (Med) & 0.25 & \textbf{0.10} & 9 & 277 & 0.12& 193 & 0.33 \\
 & High (Max) & \textbf{0.26} & 0.09 & 12 & 252 & 0.09& 208 & 0.25 \\ \midrule
\multirow{12}{*}{\rotatebox{90}{SST2}} & Zero-shot & \textbf{0.94} & \textbf{0.69} & \textbf{97} & 1297 & 0.02 & 76 & 0.77 \\
 & kNN (50) & 0.93  & 0.59  & 326 & 1758 & \textbf{0.54}& 72 & 0.62 \\
 & Low (LIAR) & 0.90 & 0.48 & 373 & 1148 & 0.15 & 87 & 0.82 \\
 & Low (Med) & 0.90 & 0.52 & 227 & \textbf{840} & 0.02 & 62 & 0.83 \\
 & Low (Max) & \textbf{0.94} & 0.58 & 252 & \textbf{784} & 0.01 & 98 & 0.81  \\
 & Mod (LIAR) & 0.93 & 0.56 & 245 & 996 & 0.20& 59 & 0.64 \\
 & Mod (Med) & \textbf{0.94} & 0.52 & 132 & 1030 & 0.04 & 34 & 0.55 \\
 & Mod (Max) & 0.93 & 0.52 & 184 & 1170 & 0.06 & 58 & \textbf{0.51}  \\
 & High (LIAR) & 0.92 & 0.62 & 365 & 1160 & 0.02 & 42 & 0.61 \\
 & High (Med) & \textbf{0.94} & 0.53 & 164 & 844 & 0.09& 52 & 0.68 \\
 & High (Max) & \textbf{0.94} & 0.61 & 320 & 857 & 0.16& 53 & 0.79 \\ \midrule
\multirow{12}{*}{\rotatebox{90}{MELD}} & Zero-shot & 0.41 & \textbf{0.15} & 39 & 808 & 0.11 & 161 & 0.77 \\
 & kNN (50) & 0.37  & 0.11  & 52 & 768 & 0.00 & 59 & 0.54 \\
 & Low (LIAR) & \textbf{0.46} & 0.14 & 20 & 532 & 0.06& 150 & 0.64 \\
 & Low (Med) & 0.45 & 0.11 & 17 & 387 & 0.30& 229 & 0.64 \\
 & Low (Max) & 0.39 & 0.09 & 30 & 477 & \textbf{0.36}& 186 & 0.65  \\
 & Mod (LIAR) & 0.40 & 0.11 & 26 & 514 & 0.00 & 106 & 0.39 \\
 & Mod (Med) & 0.40 & 0.11 & 36 & 546 & 0.03 & 130 & 0.52 \\
 & Mod (Max) & 0.38 & 0.09 & 24 & 590 & 0.08 & 110 & 0.52  \\
 & High (LIAR) & 0.44 & 0.14 & \textbf{7} & \textbf{179} & 0.09& \textbf{260} & \textbf{0.10} \\
 & High (Med) & 0.43 & 0.13 & 21 & 466 & 0.27 & 124 & 0.45 \\
 & High (Max) & 0.42 & 0.12 & 12 & 322 & 0.01 & 233 & 0.51 \\ \midrule
\multirow{12}{*}{\rotatebox{90}{AG NEWS}} & Zero-shot & 0.90 & 0.62 & 424 & 1043 & 0.00 & \textbf{108} & 0.92 \\
 & kNN (50) & 0.67  & 0.28  & 478 & 1881 & 0.09 & 22 & 0.77 \\
 & Low (LIAR) & \textbf{0.93} & \textbf{0.73} & 349 & 1183 & 0.01 & 93 & 0.90 \\
 & Low (Med) & 0.92 & 0.65 & 665 & 1589 & \textbf{0.20}& 83 & 0.89 \\
 & Low (Max) & \textbf{0.93} & 0.60 & 560 & 1164 & 0.00& 77 & 0.90 \\
 & Mod (LIAR) & 0.92 & 0.46 & 404 & 1345 & 0.12& 50 & 0.80 \\
 & Mod (Med) & \textbf{0.93} & 0.59 & 484 & 1102 & 0.01 & 48 & 0.79 \\
 & Mod (Max) & 0.92 & 0.56 & \textbf{249} & 1402 & 0.01& 57 & 0.73 \\
 & High (LIAR) & \textbf{0.93} & 0.46 & 318 & 1437 & 0.00 & 45 & \textbf{0.71} \\
 & High (Med) & \textbf{0.93} & 0.51 & 253 & \textbf{833} & 0.09& 58 & 0.86 \\
 & High (Max)  & 0.92 & 0.54 & 350 & 1576 & 0.01 & 46 & 0.73 \\ \bottomrule
\end{tabular}}
\caption{Results overview in text classification. Priors with ``(LIAR)'' suffix were calibrated during a single-objective pilot on LIAR. The same meta-parameters are then applied unchanged to every new target task. Full probability curves and all prior IDs are listed in Appendices~\ref{sec:appendix-prob-visualizations}-\ref{sec:so-experiments}.}
\label{tab:multi-objective}
\end{table}

\begin{table}[ht!]
\resizebox{1\linewidth}{!}{
\begin{tabular}{@{}p{0.30em}llllllll@{}}
\toprule
 & WS Prior & \begin{tabular}[c]{@{}l@{}}Max \\ $F1$\end{tabular} & \begin{tabular}[c]{@{}l@{}}Mean \\ $F1_m$\end{tabular} & \begin{tabular}[c]{@{}l@{}}Min \\ $ET$\end{tabular} & \begin{tabular}[c]{@{}l@{}}Mean \\$ET$ \end{tabular} & HV & \begin{tabular}[c]{@{}l@{}}No. \\ Eval\end{tabular} & \begin{tabular}[c]{@{}l@{}}Error \\ Ratio\end{tabular} \\ \midrule
\multirow{4}{*}{\rotatebox{90}{SQUAD}} & Zero-shot & 0.34 & 0.23 & 2189 & 4081 & \textbf{0.25} & \textbf{71} & 0.95 \\
 & Low (Max) & \textbf{0.89} & 0.33 & 1435 & 3150 & 0.03 & 30 & \textbf{0.76}  \\
 & Mod (Max) & 0.86 & 0.41 & 1468 & 1953 & 0.01 & 32 & 0.90  \\
 & High (Max) & \textbf{0.89} & \textbf{0.87} & \textbf{1195} & \textbf{1337} & 0.0 & 15 & 0.8 \\ \midrule
\multirow{4}{*}{\rotatebox{90}{DROP}} & Zero-shot & 0.39 & 0.18 & 2114 & 3556 & 0.11 & \textbf{96} & 0.94 \\
 & Low (Max) & 0.18 & 0.11 & 4995 & 5929 & 0.05 & 32 & 0.90  \\
 & Mod (Max) & \textbf{0.40} & 0.23 & \textbf{775} & 2259 & \textbf{0.29} & 66 & 0.86  \\
 & High (Max) & \textbf{0.40} & \textbf{0.28} & 783 & \textbf{1881} & 0.13 & 34 & \textbf{0.82} \\\bottomrule
\end{tabular}}
\caption{Results overview in Question Answering.}
\label{tab:qa}
\end{table}

Across datasets, WS priors either \emph{match} or \emph{surpass} the best Zero-shot $F1_{\text{m}}$ while systematically improving efficiency. On LIAR, a \textsc{High} prior lifts peak $F1_{\text{m}}$ from 0.24 to 0.26, cuts the mean $ET$ by a factor of 3.5, and lowers the error ratio by sevenfold. A similar pattern emerges on MELD, where \textsc{High} drives the error ratio from 0.77 to 0.10 and reduces mean $ET$ 4.5×, while keeping $F1_{\text{m}}$ above the baseline. On SST-2, the Zero-shot baseline generated the highest $F1_{\text{m}}$ and lowest $ET$ out of all variants. 

Zero‑shot runs exhibit high error ratios across all benchmarks (e.g., 0.73-0.92); the WS priors cut these failure rates dramatically, down to 0.09-0.90. Moreover, non-naive warm-started runs showed a sensible reduction in mean $ET$ while maintaining peak $F1_{\text{m}}$. On AG~News, all WS runs improve max $F1_{\text{m}}$ while several improve $ET$, HV and Error Ratio, showing that better performance–time trade-offs are discoverable even in large-scale settings.  

The naive \textbf{kNN-50} baseline, although in SST-2 case attains large HV values, degrades performance on three datasets and notably obtains the worst results out of all priors in AG NEWS (0.90 → 0.67 $F1_{\text{m}}$) and MELD (0.41 → 0.37 $F1_{\text{m}}$).

\subsection{Question Answering Results}

Table~\ref{tab:qa} reports results on the generative SQuAD~1.1 and DROP datasets.  
Knowledge reused from a single related task already yields substantial gains.  
For SQuAD, WS priors outperform the baseline in almost all metrics. The \textsc{High-Max} prior, in particular, raises $F1$ from 0.34 to 0.89 while shrinking mean ET from 4081s to 1337s (-3×). Similarly to the text classification results, WS priors bring error ratios down from 0.94–0.95 (zero‑shot) to 0.76–0.90.

On DROP, the \textsc{Low} prior illustrates negative transfer, yet both \textsc{Moderate} and \textsc{High} priors outperform Zero-shot on \emph{every} metric; peak $F1$ improves slightly (0.39→0.40) and mean ET falls by 47\%. These outcomes confirm that cross-task meta-knowledge generalizes beyond classification and that the adaptive pull/push schedule mitigates catastrophic transfers.

\section{Discussion} \label{sec:discussion}

\begin{figure}[h]
\includegraphics[width=\linewidth]{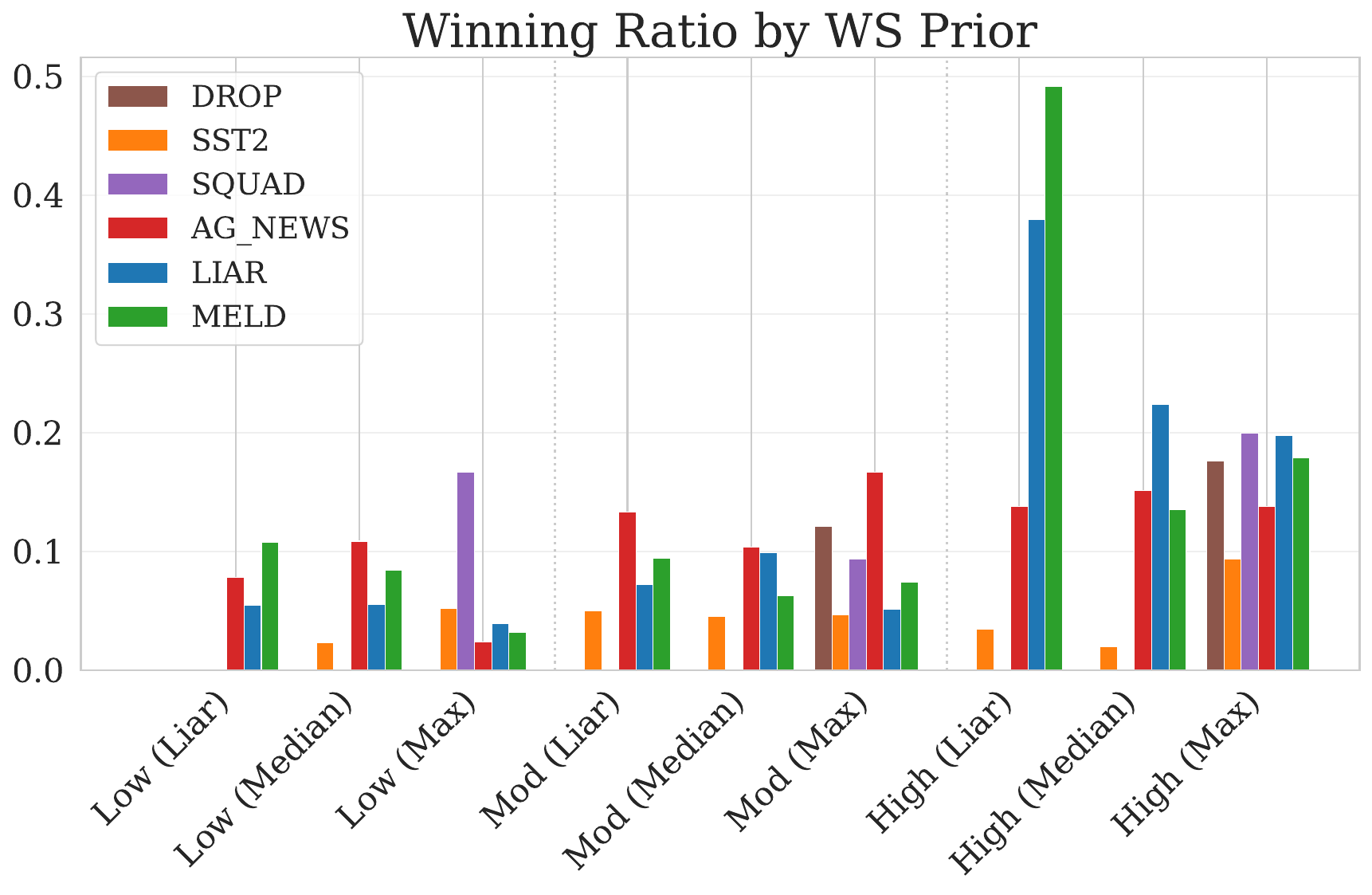}
    \caption {Ratio of discovered pipelines outperforming the Zero-shot baseline in Text Classification and QA.}
    \label{fig:wins-by-configs}
\end{figure}

Warm-start priors consistently steer the search toward stronger performance–time trade-offs across all six benchmarks. Figure~\ref{fig:wins-by-configs} reports the winning ratio: the share of evaluated LM pipelines that improve upon the zero-shot Pareto front. 

\begin{figure*}[t]
\centering
\begin{subfigure}{0.47\linewidth}
    \includegraphics[width=\linewidth]{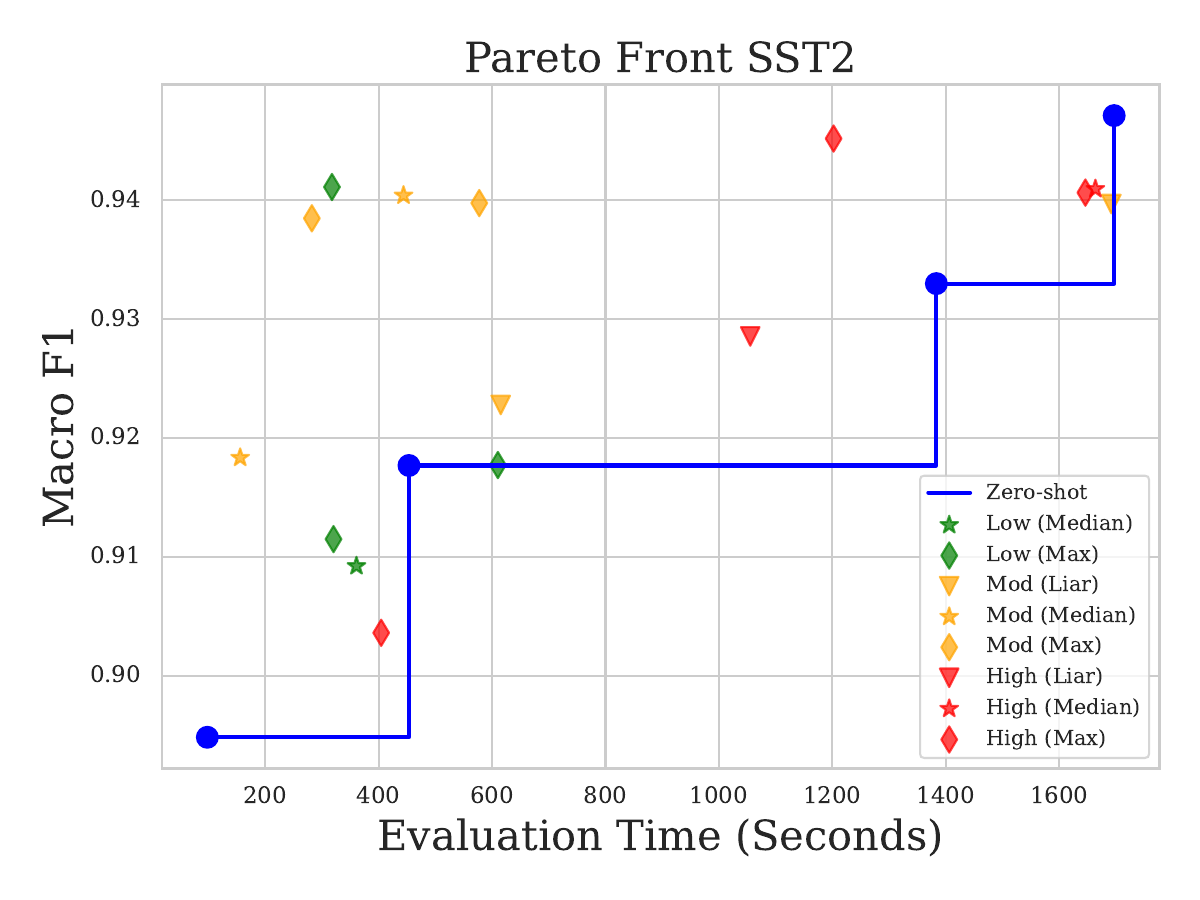}
    \caption{}
    \label{fig:pareto-sst2-1}
\end{subfigure}
   \hfill
\begin{subfigure}{0.47\linewidth}
  \includegraphics[width=\linewidth]{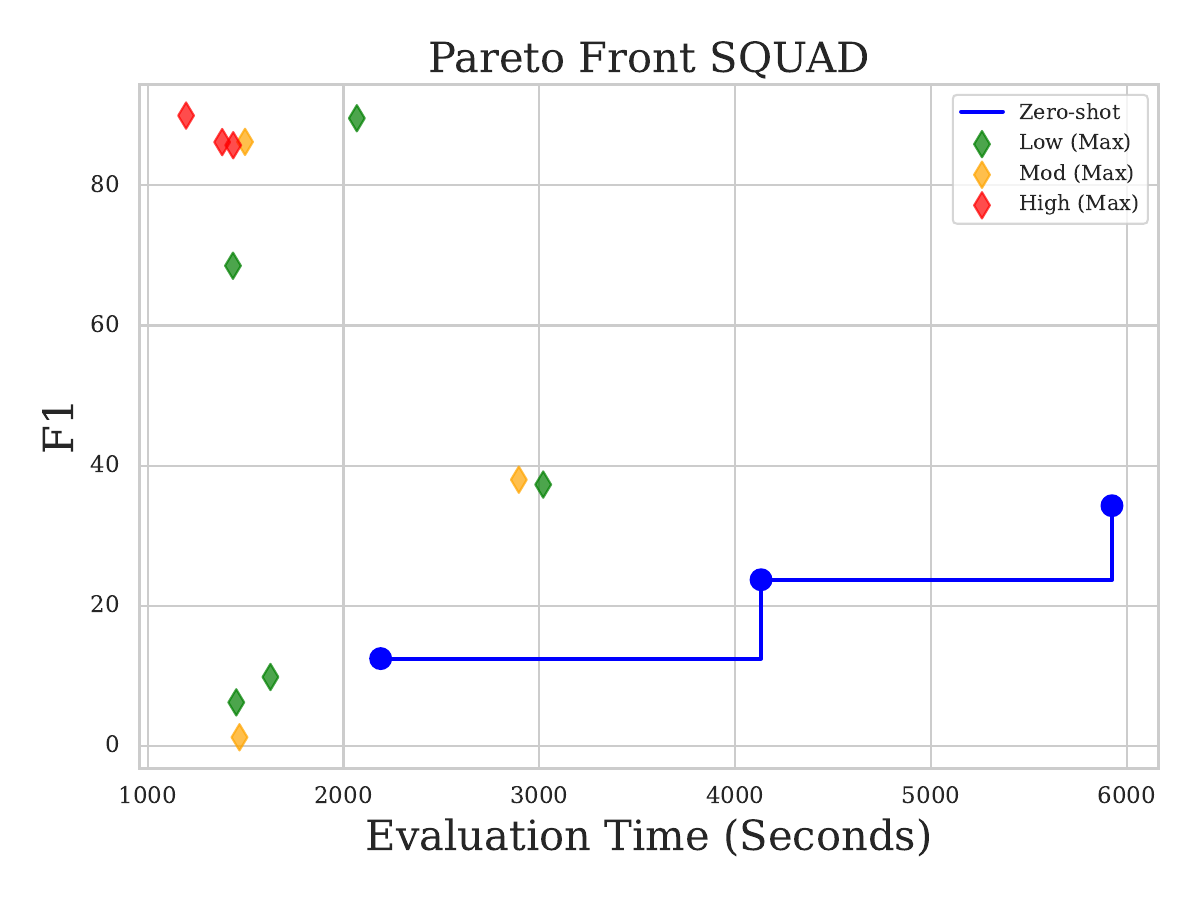}
    \caption{}
    \label{fig:pareto-squad-1}
\end{subfigure}

  \caption {Pareto Fronts discovered by the different Priors on SST2 (a) and SQUAD (b).}
  \label{fig:pareto-1}
\end{figure*}

The \textsc{High–Max} prior is the most stable, winning about 20\% of pipelines on SQuAD, LIAR, MELD, and DROP, and 10–15\% on SST-2 and AG News. On the LIAR and MELD pair, the \textsc{High–LIAR} prior achieves winning ratios near 50\% and 40\%, respectively, while cutting the error rate by a factor of seven (Table~\ref{tab:multi-objective}). For clarity, all ET values are computed only on successful evaluations, while failure rates are captured by the Error Ratio, with all methods allotted an identical 24 GPU‑hour wall‑clock budget per run.

\begin{figure}[ht!]
\centering
\includegraphics[width=\linewidth]{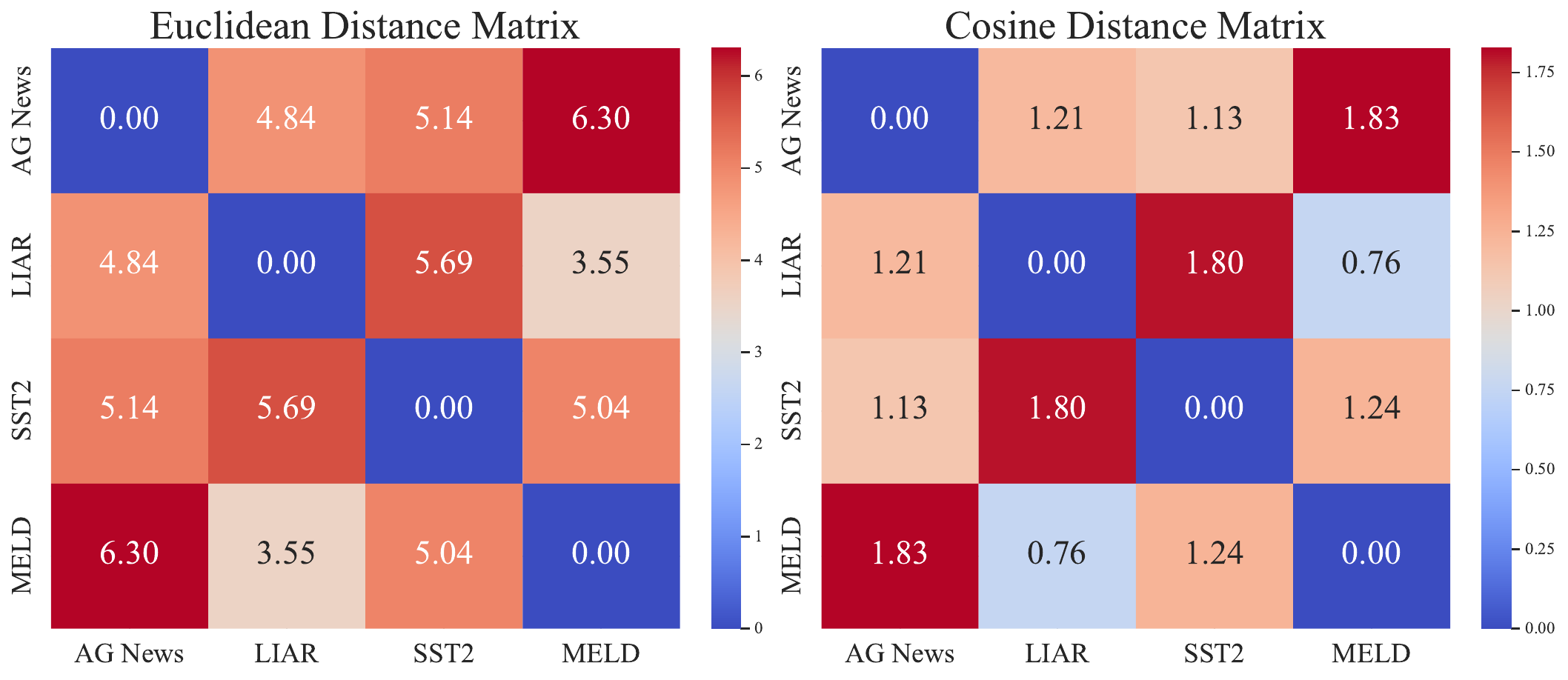}
    \caption {Distance between Text Classification Tasks according to their meta-features (Section \ref{sec:meta-features}).}
    \label{fig:distance-heatmaps}
\end{figure}

These results show that combining experience discrimination with adaptive probability shifts yields the best of both worlds: rapid convergence when relevant meta-knowledge exists yet robustness when it does not. Whenever the experience store contained closely related traces, e.g., MELD–LIAR (Figure~\ref{fig:distance-heatmaps}), the similarity-aware priors trimmed average evaluation time by up to 4.5x and increased peak $F_{1_{\text{m}}}$ (Table~\ref{tab:multi-objective}). Even on sparsely related tasks such as SST-2 and AG News, softer pulls uncovered superior Pareto trade-offs by moderating exploration strength (Figure~\ref{fig:pareto-sst2-1}).

The baseline performance of kNN highlights the significance of selective memory. While it has access to both positive and negative examples, it assigns equal weight to all neighbors, failing to demote weak configurations and causing accuracy to fall on three of four classification datasets. In contrast, XAutoLM’s asymmetric pull–push update penalizes both past failures and underperforming successes. DROP, for example, illustrates the need to learn from failures: a low-bias prior that ignores negatives collapses to $F_{1}=0.18$, whereas reinstating the push restores $F_{1}=0.40$ and halves mean evaluation time.

Our findings further show that transfer using our method extends beyond classification. With barely a handful of relevant experience, a high-bias prior multiplies SQuAD $F1$ from $\approx0.3$ to $\approx0.9$ and compresses evaluation time by threefold, producing a dominant Pareto front (Figure~\ref{fig:pareto-squad-1}). On the other hand, DROP illustrates the importance of negative experiences: a low-bias prior that ignores negatives collapses to $F_{1}=0.18$, whereas reinstating the push restores $F_{1}=0.40$ and cuts mean evaluation time by 50 \% (Table~\ref{tab:qa}).

A core motivation of our framework is to reduce the carbon footprint and environmental toll of repeated large-scale language model fine-tuning. By systematically reusing insights from past runs, \textbf{XAutoLM} significantly reduces redundant evaluations and lowers the overall error rate during the search. Beyond simply lowering compute hours, this approach aligns with the growing Green AI ethos in NLP \citep{wang2023energy,schwartz2020green}, emphasizing the importance of responsible resource usage. Our experiments demonstrate that our warm-start strategy enhances performance and streamlines the search process, resulting in algorithms that strike a better balance between efficiency and performance.


\section{Conclusions} \label{sec:conclusions}

XAutoLM converts the costly trial–and–error of language model fine-tuning into a guided, resource-aware search.  By seeding the optimizer with a similarity-weighted prior built from past \emph{successes \& failures}, the framework consistently uncovers pipelines with superior performance–time trade-offs. Across four text-classification corpora and two generative QA benchmarks, it surpasses the best zero-shot $F_{1}$ on five tasks, matching it on SST-2, while cutting mean pipeline evaluation time by \emph{up to} a factor of four and reducing error rates by \emph{as much as} sevenfold. These gains hold across a refreshed model pool that ranges from lightweight discriminative to compact generative models. Because every recovered pipeline reuses information already paid for, XAutoLM advances the \emph{Green AI} agenda \citep{schwartz2020green}, delivering competitive results in less search time, while avoiding redundant computation.

\section{Limitations} \label{sec:limitations}

We identify some limitations to our study that highlight avenues for further investigation:

\subsubsection*{Scaling to bigger LLMs}

XAutoLM is \textit{scale-agnostic}: the optimizer treats candidates as black-box fit/evaluate calls and does not rely on model internals. Our open-source implementation presently evaluates on a \textbf{single GPU}, which constrained the largest models tested; this is a property of the evaluator backend, not of the optimization method. The experience store logs a minimal hardware profile (Section \ref{sec:exp-store}), which helps steer the search away from infeasible pipelines under a given machine with a single GPU setup. Supporting larger models, therefore, amounts to adding multi-GPU meta-features and swapping in a larger-model evaluator (e.g., parameter-efficient  \cite{lora}/quantized \cite{nagel2021white,dettmers2023qlora} or distributed evaluators \cite{zhao2023pytorch}) in future releases; the search algorithm and experience-based priors remain unchanged. We leave such engineering backends to future work and keep our claims limited to the single-GPU setting evaluated here.

\subsubsection*{Multimodality}
The current experience store and benchmarks are text-only; verifying that the warm-start prior transfers to dialogue, speech, or multimodal pipelines is an essential next step.

\subsubsection*{Statistical Tests}
Statistical support is available only for the single-objective probes archived in Appendix~\ref{sec:so-experiments}. Extending significance testing to the multi-objective fronts of Tables~\ref{tab:multi-objective} and \ref{tab:qa} would require many repeated runs and is left for future work, where bootstrap or fully Bayesian analyses are planned.

\subsubsection*{Efficiency Measures}
Our energy discussion rests on the empirical link between execution time and power draw reported by prior work \cite{wang2023energy,estevanell2024balancing}; we did not log wattage directly. The next release of XAutoLM will record real-time power and emit CO\textsubscript{2} estimates alongside performance metrics.

\section*{Acknowledgments}
This research has been partially funded by the University of Alicante, the University of Havana, the Spanish Ministry of Science and Innovation, the Generalitat Valenciana, and the European Regional Development Fund (ERDF) through the following funding: At the regional level, and as the primary source of support, the Generalitat Valenciana (Conselleria d'Educacio, Investigacio, Cultura i Esport), FEDER granted funding for CIDEGENT (CIDEXG/2023/13); and NL4DISMIS (CIPROM/2021/21). At the national level, the following projects were granted:  HEART-NLP (PID2024-156263OB-C22); COOLANG (PID2021-122263OB-C22); SOCIALTRUST (PDC2022-133146-C22); ILENIA (2022/TL22/00215334) and ALIA models (\url{https://alia.gob.es}) funded by {MCIN/AEI/10.13039/501100011033} and, as appropriate, by ERDF A way of making Europe, by the European Union or by the European Union NextGenerationEU/PRTR; and by the State Subprogram for Training, Attraction, and Retention of Talent (PEICTI 2024) of the Spanish Ministry of Science and Innovation, grant PRX24/00272.

\bibliography{acl_latex}
\appendix

\section{Additional Implementation Details and Experimental Configurations}
\label{sec:appendix-implementation-details}

In this section, we provide key implementation details to ensure that our work is fully reproducible. All configuration candidates used in our multi-objective and single-objective experiments are available in Appendix~\ref{sec:appendix-prob-visualizations} and Appendix~\ref{sec:so-experiments} due to the extremely high number of tested configurations. In our evaluations, candidate configurations were designed with two distinct learning rate schemes and distance discrimination strategies, as detailed below.

\subsection{Learning Rate Configuration and Update Strategy}

We adopt a dual-mode configuration for the learning rate updates applied to the probabilistic model. In experiments employing fixed learning rates, we set the parameters to 
\[
\alpha_{\max}^{+} = 0.05 \quad \text{and} \quad \alpha_{\max}^{-} = -0.02.
\]
For configurations using adaptive learning rates, the values are computed as 
\[
\alpha_{\max}^{+} = \frac{1}{N_{\text{pos}}} \quad \text{and} \quad \alpha_{\max}^{-} = -\frac{1}{N_{\text{neg}}}
\]
Where \(N_{\text{pos}}\) and \(N_{\text{neg}}\) denote the number of positive and negative experiences, respectively. Although these rates are expressed with positive and negative signs to indicate the direction of the update (reinforcing or de-emphasizing a configuration), all update steps are executed using the absolute values.

\subsection{Normalization of Meta-Features}

All meta-features used for computing distances are standardized using a standard scaler normalizer. This normalizer computes the mean and standard deviation of the feature vectors (with a small epsilon added to avoid division by zero) and returns the standardized data. This ensures that distance computations are robust and comparable across features.

\subsection{Beta Scale and Utility Functions}

For the decay parameter \(\beta\), two formulations are employed: the \emph{std-only} beta scale is used in single-objective experiments, whereas the \emph{std-plus-mean} beta scale is applied in multi-objective settings.

All candidates for the single-objective experiments (Appendix \ref{sec:so-experiments}) utilize a weighted sum approach with the \(F1\) score weight set to 1 and the evaluation time weight set to 0. Detailed specifications of candidate configurations can be found in the visualizations provided in the respective sections (Appendix \ref{sec:so-experiments} for single-objective, and Appendix \ref{sec:appendix-prob-visualizations} for multi-objective).

\subsection{Experimental Setup and Computational Resources}

The main text fully discloses our experimental setup (Section \ref{sec:experiments}).

\subsection{Framework Overview and Dependencies}
XAutoLM is implemented on top of the AutoGOAL framework~\cite{estevanell2024balancing,estevez2020automatic}, leveraging its optimization strategy and abstractions. Our implementation is developed in Python and utilizes the HuggingFace Transformers library \cite{wolf2019huggingface} to access pre-trained language models. A complete list of dependencies, environment setup instructions, and detailed documentation on how to run the experiments (and statistical testing), reproduce the results, and navigate the codebase is provided in the repository.

The code and all associated materials can be accessed at the following GitHub repository: \url{https://github.com/EEstevanell/XAutoLM}. 

\section{Multi-Objective Initial Probabilities}
\label{sec:appendix-prob-visualizations}

This appendix visualizes the initial probability distributions over fine-tuning methods induced by different meta-learning configurations (Prior) in our multi-objective experiments (see Section\ref{sec:experiments}). Each configuration is defined by:
\begin{enumerate}
\item Inclusion of positive and/or negative experiences,
\item Utility function (Weighted Sum, Linear Front, Logarithmic Front),
\item Distance metric (Euclidean, Cosine) with scaling, and
\item Pull/push limits $k_{\mathrm{pos}}$, $k_{\mathrm{neg}}$ and learning-rate scheme (fixed/adaptive).
\end{enumerate}

Recall that we generated up to 180 candidate configurations per dataset by systematically varying:
\begin{enumerate}
    \item Inclusion/exclusion of \emph{positive} (successful) and \emph{negative} (error) past experiences,
    \item Utility functions (e.g., weighted sum, linear front, logarithmic front),
    \item Distance metrics (Euclidean, Cosine) and their scaling,
    \item \(\alpha_{\max}^{+}\) and \(\alpha_{\max}^{-}\) values (fixed or adaptive) (Section \ref{sec:warmstart}).
\end{enumerate}
Each configuration yields a distinct initial probability vector for the available fine-tuning methods, with deviations from the baseline distribution measured via \textit{Total Variation} (TV). Grouping configurations by TV allows us to categorize them into \emph{low}, \emph{moderate}, and \emph{high} bias levels relative to the baseline's uniform initialisation.

\subsection{Classification Tasks}
\begin{figure*}[ht!]
\includegraphics[width=\linewidth]{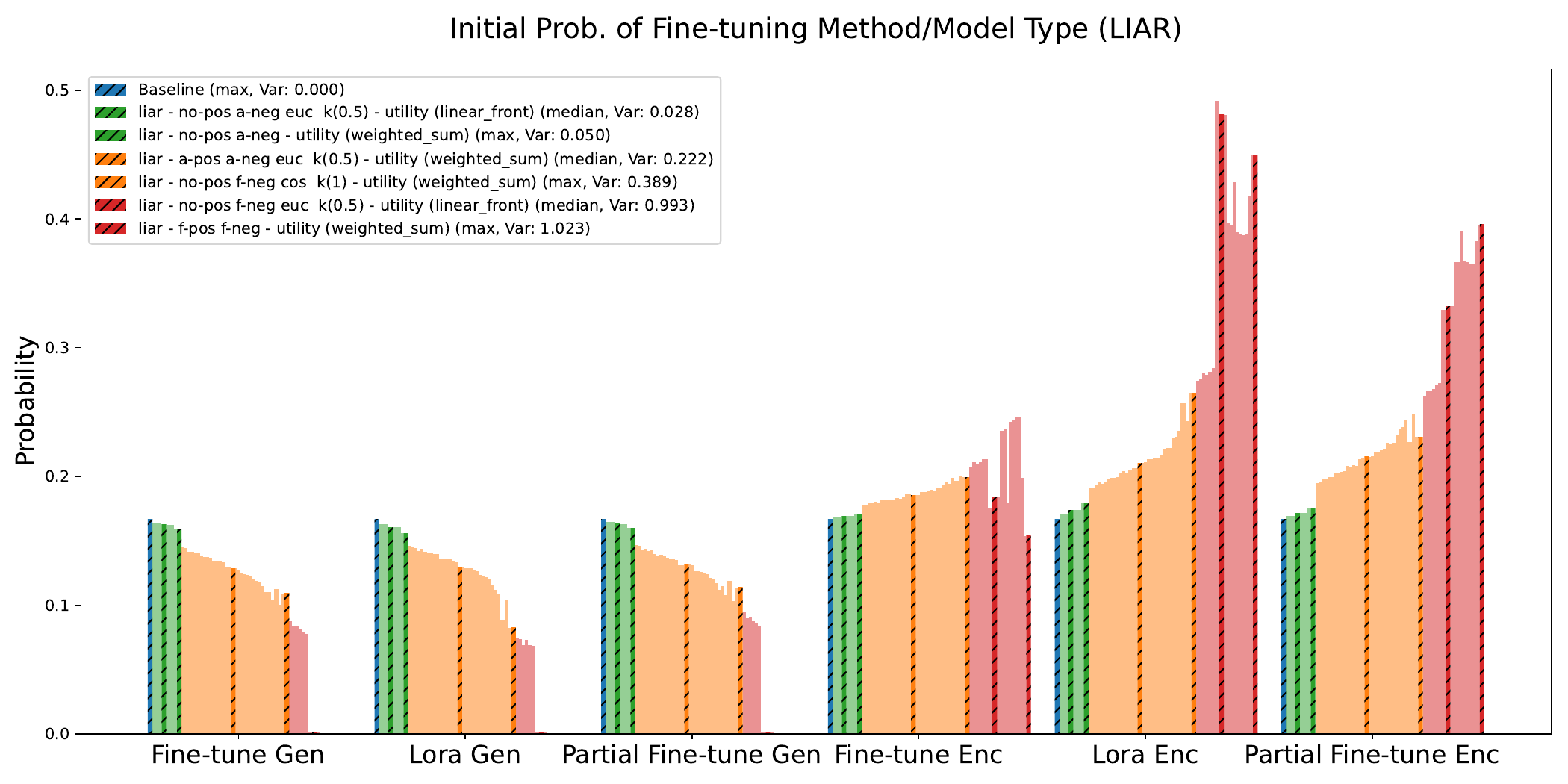}
    \caption {Initial probability distributions for fine-tuning methods on LIAR.}
    \label{fig:initial-probs-liar}
\end{figure*}

\begin{figure*}[ht!]
\includegraphics[width=\linewidth]{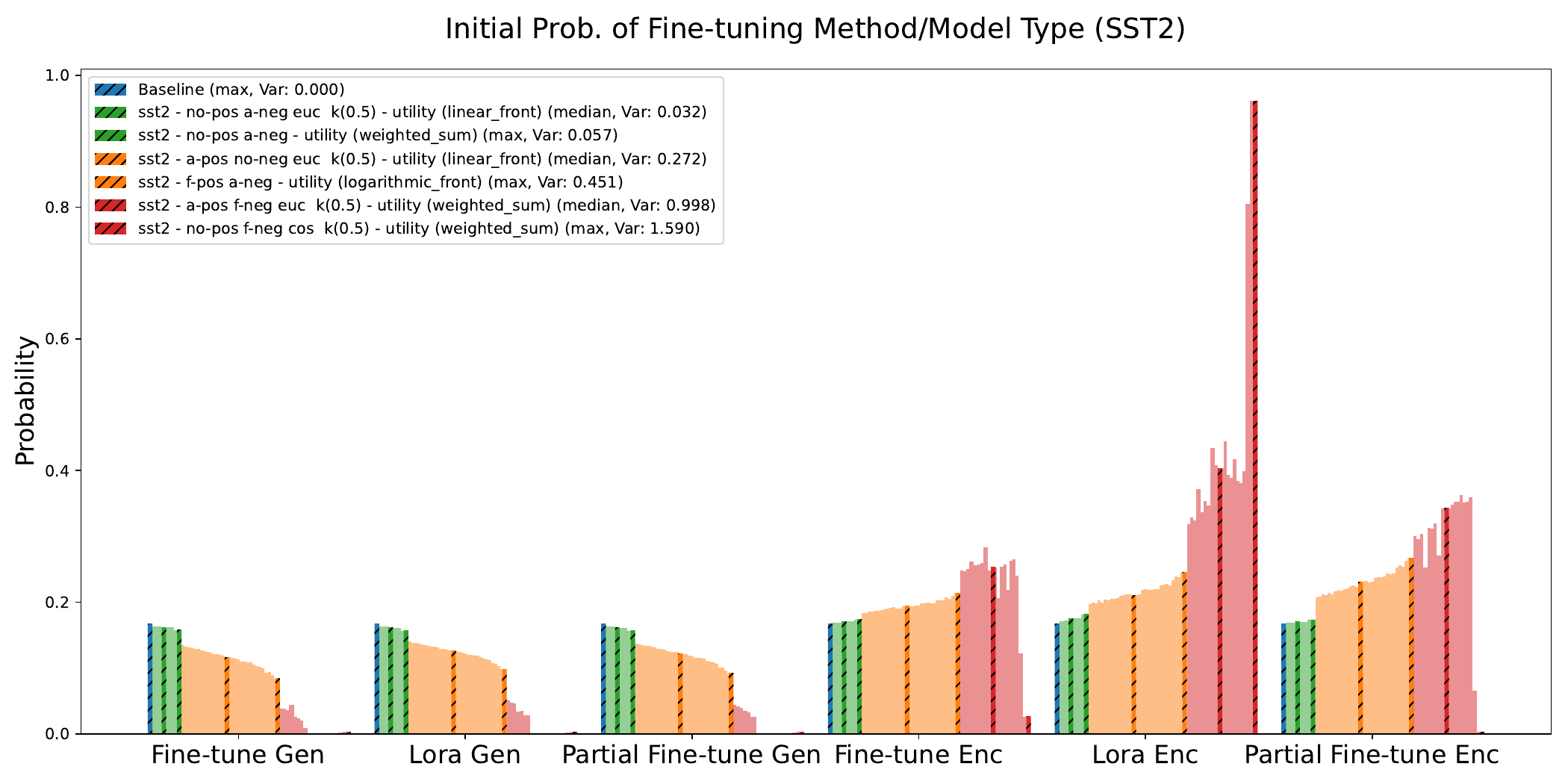}
    \caption {Initial probability distributions for fine-tuning methods on SST2}
    \label{fig:initial-probs-sst2}
\end{figure*}

\begin{figure*}[ht!]
\includegraphics[width=\linewidth]{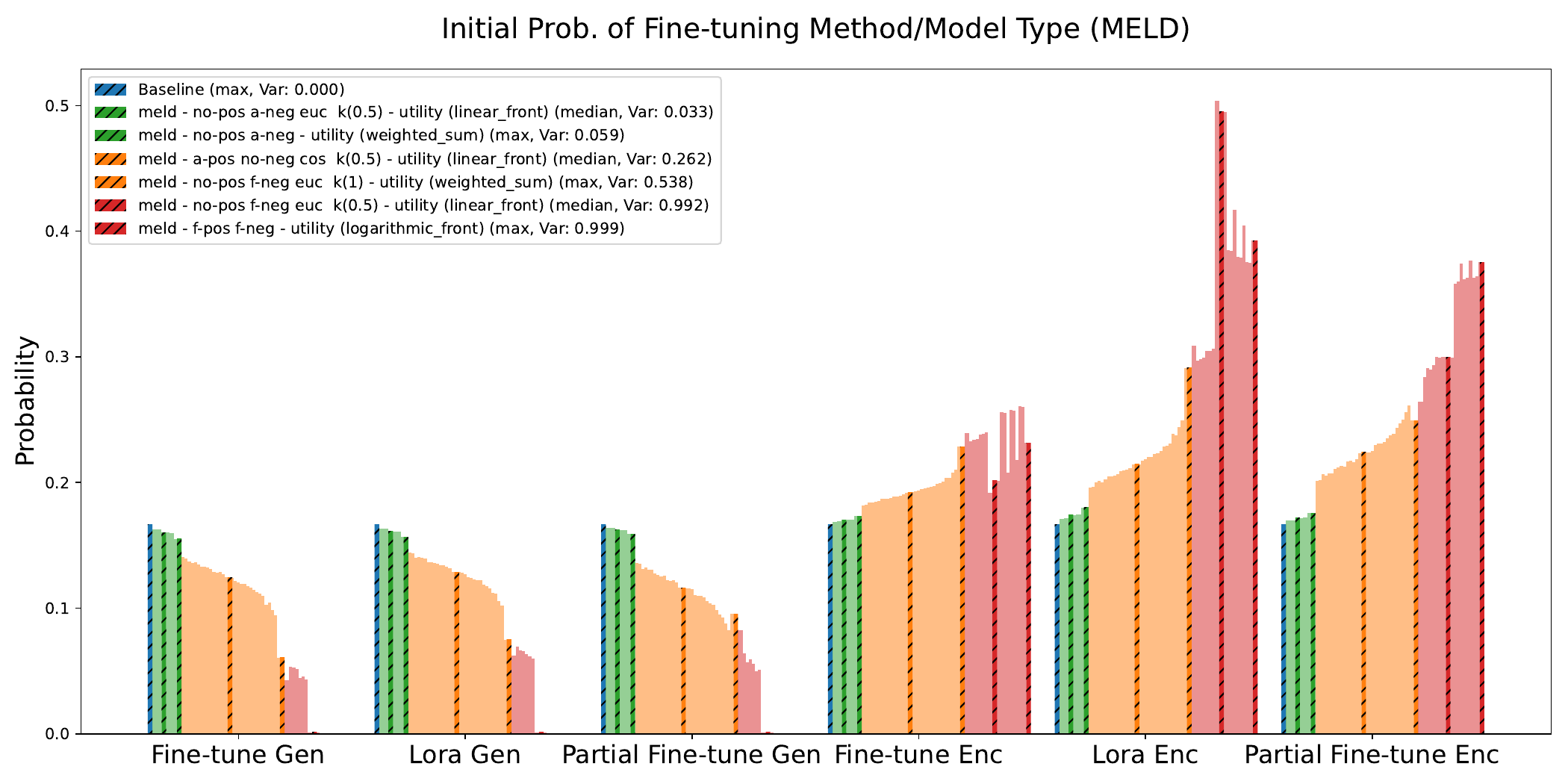}
    \caption {Initial probability distributions for fine-tuning methods on MELD}
    \label{fig:initial-probs-meld}
\end{figure*}

\begin{figure*}[ht!]
\includegraphics[width=\linewidth]{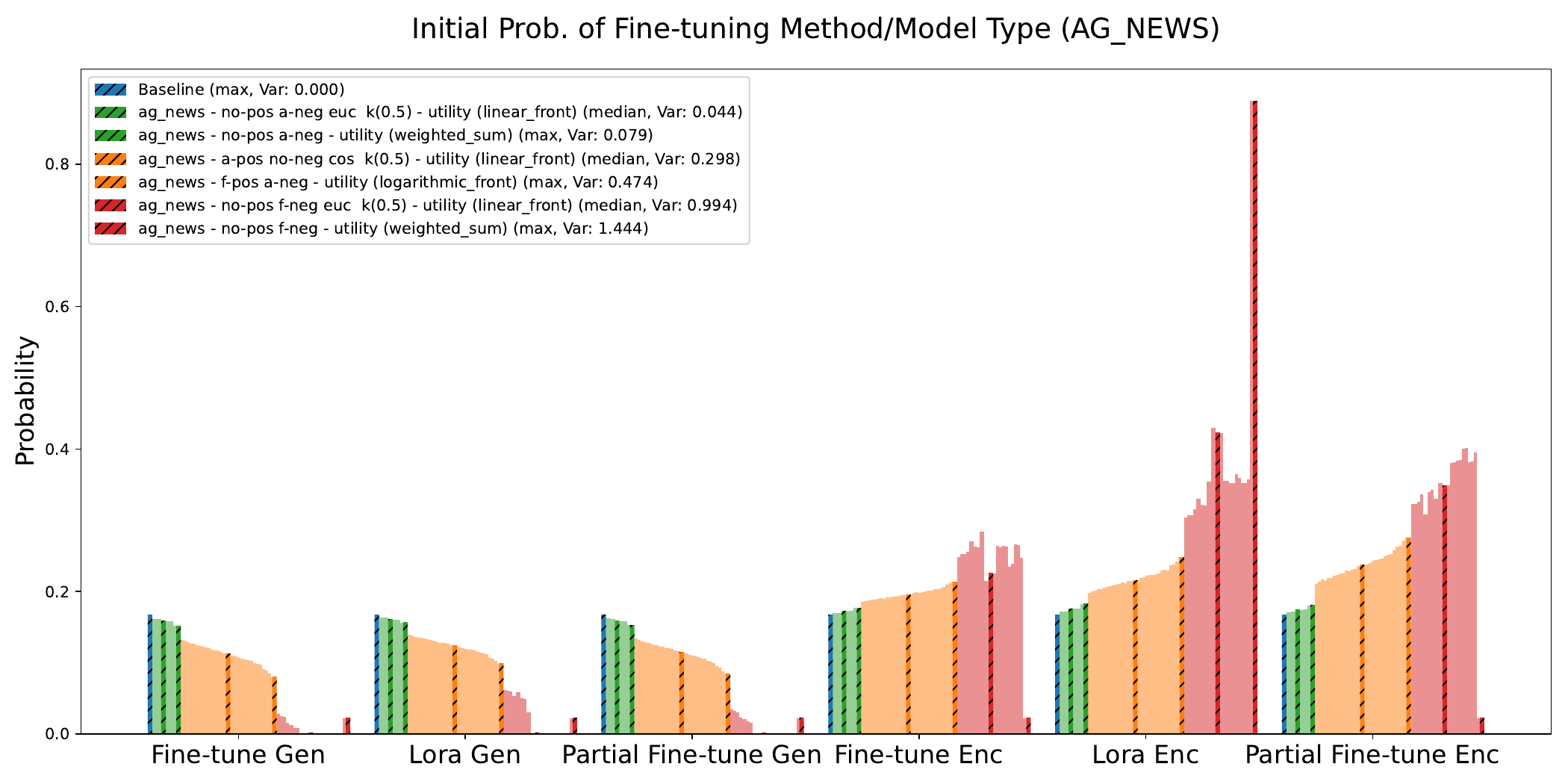}
    \caption {Initial probability distributions for fine-tuning methods on AG News}
    \label{fig:initial-probs-ag_news}
\end{figure*}

For each classification dataset (LIAR, SST-2, MELD, AG News), Figures~\ref{fig:initial-probs-liar}--\ref{fig:initial-probs-ag_news} plot the initial probabilities for representative configurations at each bias level. In each figure:
\begin{itemize}
\item \textbf{Blue:} Uniform baseline.
\item \textbf{Green, Orange, Red:} Increasing TV distance (Low, Moderate, High).
\item \textbf{Patterned Bars:} Selected \emph{Max-TV} configuration within each bin.
\end{itemize}

\paragraph{LIAR.}
Figure~\ref{fig:initial-probs-liar} shows the initial probabilities of using each fine-tuning method for the \textsc{liar} dataset, sorted by their overall difference from the baseline. Blue bars indicate the baseline configuration, whereas green, orange, and red bars represent configurations increasingly diverging from the baseline. We marked selected \emph{representative} configurations (patterned bars) for each bias level.

\paragraph{SST2.}
Figure~\ref{fig:initial-probs-sst2} illustrates the same analysis on \textsc{sst2}. Although the dataset differs substantially from \textsc{liar} regarding meta-features (e.g., number of classes, data size, label distribution), we observe a similar pattern in how the bias level shifts probabilities among alternative fine-tuning methods. The High (Max) configuration notably shows more aggressiveness than LIAR's.

\paragraph{MELD.}
Figure~\ref{fig:initial-probs-meld} shows the \textsc{meld} dataset’s initial distributions. As discussed in Section~\ref{sec:experiments}, \textsc{meld} shares some meta-feature similarities with \textsc{liar} (Figure~\ref{fig:distance-heatmaps}), causing some distributions to concentrate around methods found promising in \textsc{liar}’s prior runs.

\paragraph{AG News.}
Lastly, Figure~\ref{fig:initial-probs-ag_news} displays the candidate configurations for \textsc{ag news}, a large corpus with four news categories. 

\subsection{QA Tasks}
\begin{figure*}[ht!]
\begin{subfigure}{0.5\linewidth}
  \includegraphics[width=\linewidth]{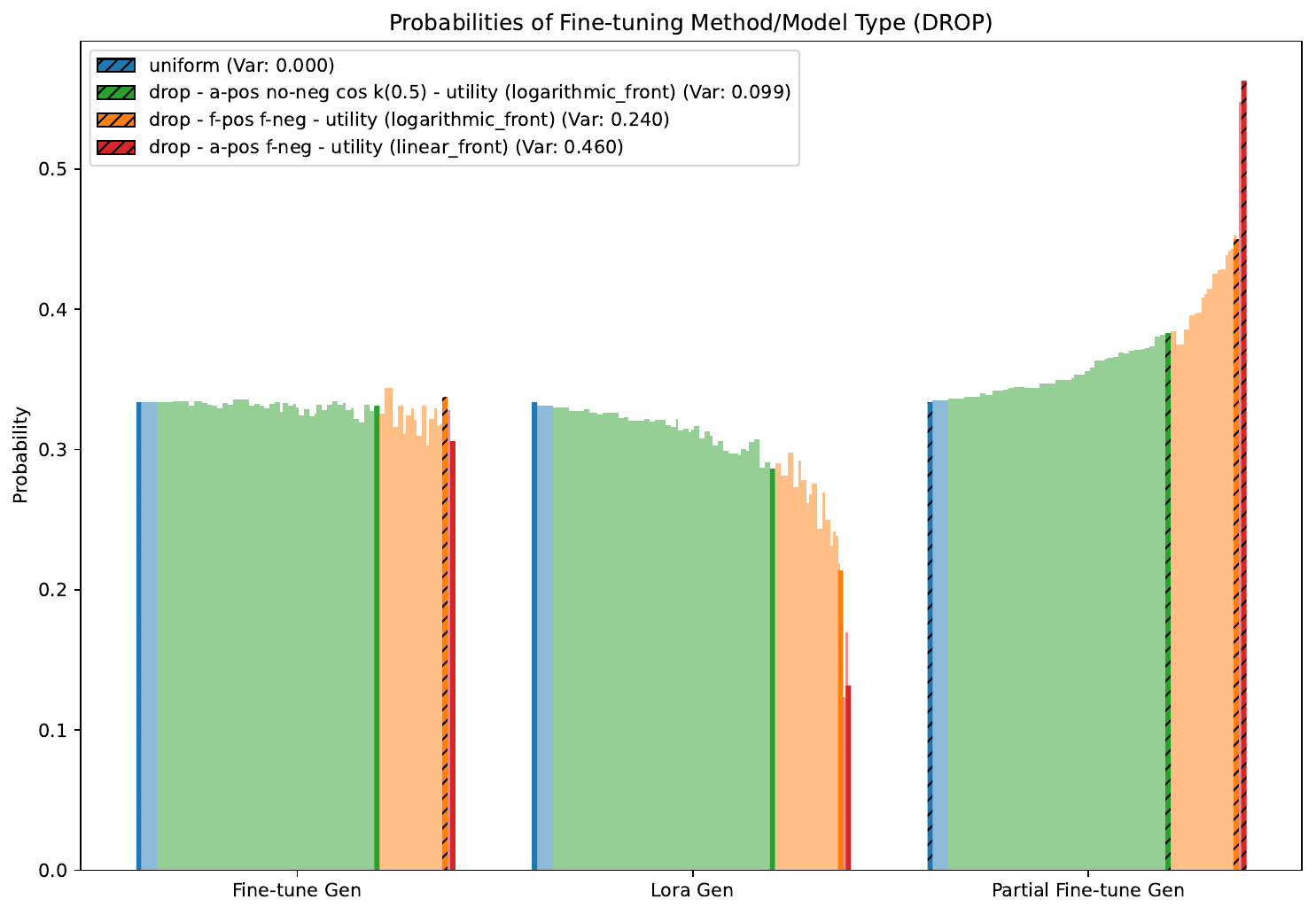}
    \caption{}
    \label{fig:initial-probs-drop}
\end{subfigure}
   \hfill
\begin{subfigure}{0.5\linewidth}
  \includegraphics[width=\linewidth]{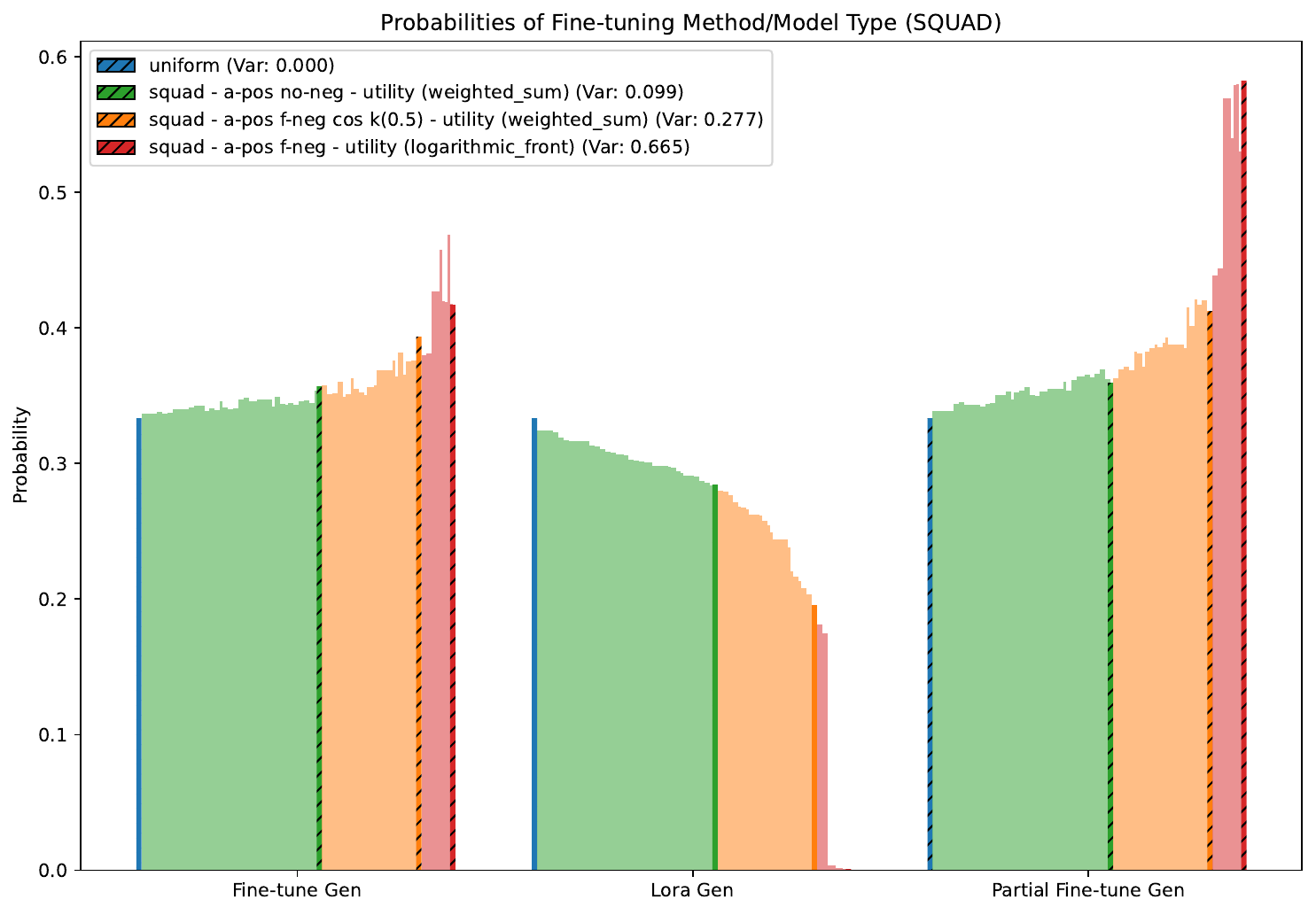}
    \caption{}
    \label{fig:initial-probs-squad}
\end{subfigure}
    \caption {Initial probability distributions for fine-tuning methods on DROP (a) and SQUAD (b)}
    \label{fig:initial-probs-drop-squad}
\end{figure*}

Figures~\ref{fig:initial-probs-drop} and \ref{fig:initial-probs-squad} show the analogous distributions for DROP and SQuAD. Despite fewer experiences, meta-learning concentrates probability mass on the partial and traditional fine-tuning strategy while avoiding Lora.

These visualizations underscore how our meta-learning strategy adapts the search space before optimization begins. By systematically adjusting the initial probabilities, XAutoLM avoids mindlessly searching all possibilities and exploits task similarities to emphasize configurations that are historically more successful or resource-feasible.

\newpage\  
\newpage\ 
\newpage\ 

\section{Single-Objective Warm Start Evaluation}  \label{sec:so-experiments}

This appendix reports single-objective experiments optimizing the macro-$F1$ score alone. We compare the Zero-shot AutoGOAL baseline against three representative warm-start priors, Low, Moderate, and High bias, selected from fourteen candidate configurations grouped by total variation (TV) distance. All priors use the std-only $\beta$ scale, Euclidean distance, and fixed learning rates ($\alpha_{\max}^+=0.05$, $\alpha_{\max}^-=0.02$).

\subsection{Initial Probability Distributions}
\begin{figure*}
    \centering
    \includegraphics[width=0.90\linewidth]{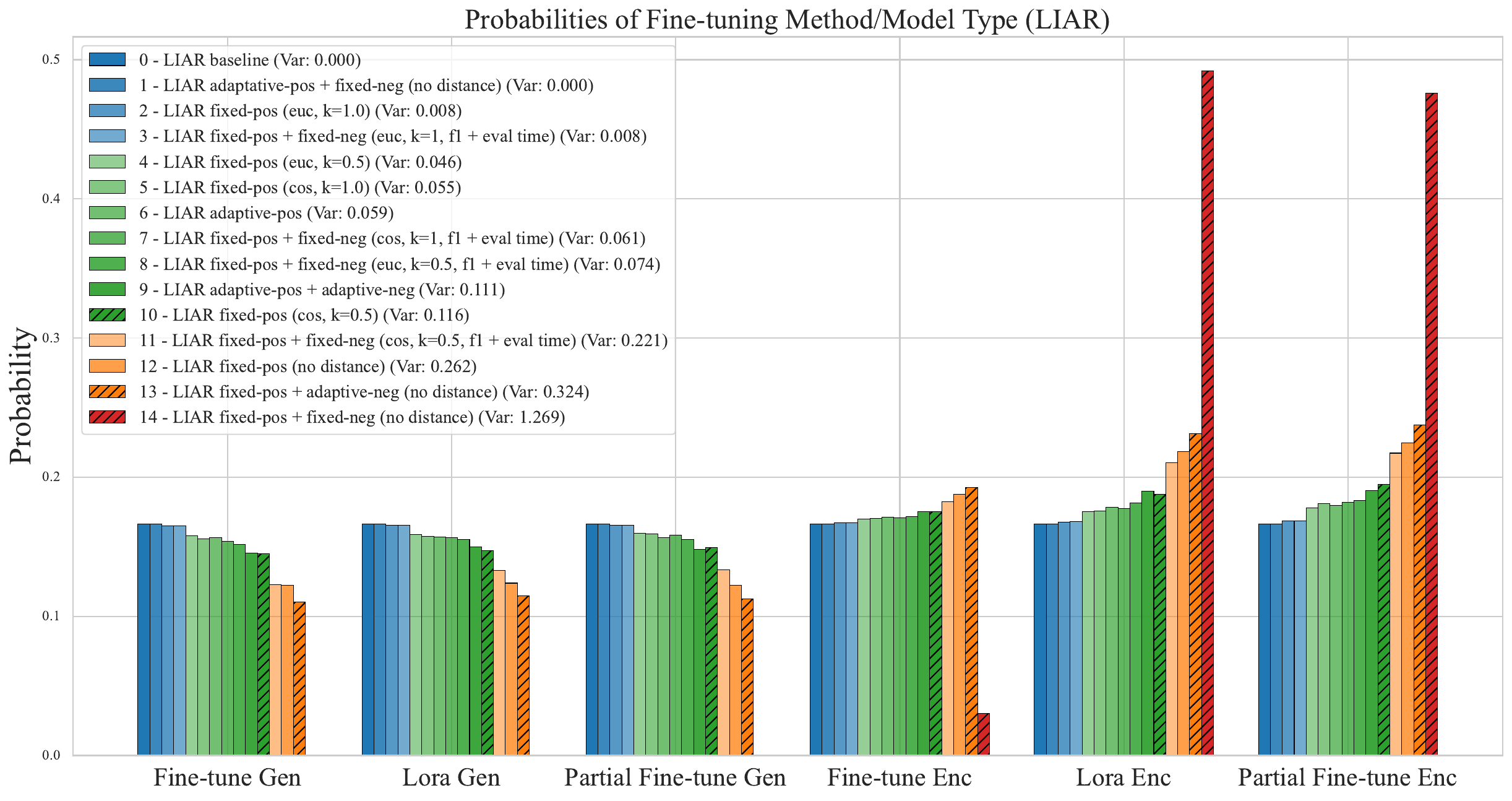}
    \caption{Initial fine-tuning probabilities for LIAR under fourteen priors, sorted by TV. Solid blue denotes the uniform baseline; patterned green, orange, and red denote the Low, Moderate, and High bias priors, respectively}
    \label{fig:prob-liar}
\end{figure*}

Figure~\ref{fig:prob-liar} shows LIAR's initial fine-tuning method distributions under the fourteen meta-learning priors, sorted by TV relative to the uniform baseline. The solid blue bar indicates the baseline; patterned green, orange, and red bars mark the chosen Low, Moderate, and High priors.

\subsection{Performance Results}

Table~\ref{tab:single-objective} reports our results. We conducted a detailed statistical analysis across six independent runs per configuration on LIAR and SST-2, evaluating performance, convergence time, and reliability. Normality was tested using Shapiro–Wilk, followed by ANOVA \cite{mchugh2011multiple} for normal metrics, and Friedman tests \citep{pereira2015overview} for non-parametric ones. We report Cohen’s $d$ and Cliff’s $\delta$ as effect‐size measures; power analyses accompany each test in the repository.

\begin{table*}[h!]
\centering
\resizebox{0.9\linewidth}{!}{
\begin{tabular}{@{}lllllllll@{}}
\toprule
Dataset & Config. & Max $F1_{m}$ & Mean $F1_{m}$ & $TT50$ (h) & $TT75$ (h) & $TT90$ (h) & No. Eval & E. Ratio \\ \midrule
\multirow{4}{*}{LIAR} & Baseline & 0.248 ±0.018 & 0.09 ±0.004 & 2.00 & 6.38 & 8.15 & \textbf{173} & 0.69 \\
 & Low WS & \textbf{0.253 ±0.006} & \textbf{0.11 ±0.008} & \textbf{1.35} & \textbf{4.10} & 9.05 & 166 & 0.61 \\
 & Mod WS & 0.251 ±0.015 & \textbf{0.11 ±0.008} & 1.57 & 4.88 & \textbf{6.43} & 165 & 0.46 \\
 & High WS & 0.247 ±0.006 & 0.10 ±0.009 & 1.37 & 5.42 & 10.74 & 156 & \textbf{0.24} \\ \midrule
\multirow{4}{*}{SST2} & Baseline & 0.928 ±0.018 & 0.56 ±0.053 & 1.69 & 2.07 & 4.64 & 85 & 0.83 \\
 & Low WS & 0.917 ±0.016 & \textbf{0.59 ±0.063} & 1.28 & 2.41 & 5.09 & \textbf{98} & 0.80 \\
 & Mod WS & \textbf{0.941 ±0.004} & 0.56 ±0.064 & 0.70 & 3.88 & 5.21 & 55 & 0.69 \\
 & High WS & 0.932 ±0.002 & 0.56 ±0.058 & \textbf{0.41} & \textbf{0.41} & \textbf{2.23} & 58 & \textbf{0.58} \\ \bottomrule
\end{tabular}
}
\caption{Overview of XAutoLM performance on optimising $F1_{macro}$ for LIAR and SST2. Results are averaged over six runs with different seeds. ‘Max $F1_m$’ and ‘Mean $F1_m$’ show the mean and standard deviation, respectively; ‘TT50’, ‘TT75’, and ‘TT90’ report the average time to reach 50\%, 75\%, and 90\% $F1_m$; and ‘No. Eval’ and ‘E. Ratio’ indicates the average number of pipeline evaluations and the ratio of such evaluations that were errors.}
\label{tab:single-objective}
\end{table*}

On \textbf{LIAR}, while none of the warm-start priors significantly outperformed the baseline in peak $F1_{\text{macro}}$ (ANOVA $p=0.856$, Friedman $p=0.94$), we observed a significant overall improvement in \emph{mean} performance across groups (ANOVA $p=0.005$, Friedman $p=0.004$). Post-hoc comparisons, however, were not significant after correction, likely due to limited sample size. More notably, the \emph{error ratio}, the share of failed evaluations, dropped dramatically from 0.69 (baseline) to 0.24 (High WS), a difference found to be statistically significant (Friedman $p=0.031$) with a large effect size (Cohen’s $d=3.39$). Convergence time metrics (TT50, TT75, TT90) also trended lower, with moderate effect sizes, although these differences did not reach statistical significance.

On \textbf{SST-2}, the Mod WS prior achieved the highest max $F1_{\text{macro}}$ (0.941), and the ANOVA test confirmed a significant group effect ($p=0.031$). The error ratio again showed a significant overall effect (Friedman $p=0.038$), improving from 0.83 (baseline) to 0.58 (High WS). Convergence time reductions were most pronounced with the High WS prior, which reached 50\% of peak $F1$ four times faster than the baseline (0.41h vs.\ 1.69h). While these improvements showed large effect sizes (e.g., TT50 $d=0.55$), they were not statistically significant in pairwise tests, most likely due to low sample power ($n=6$).

In summary, warm-start priors consistently yielded practical convergence speed and robustness benefits. While not all improvements were statistically significant, expected under a small-sample regime, our analysis shows that key metrics such as error ratio and mean F1 on LIAR and max F1 on SST-2 do reach significance. Full results, post hoc comparisons, and power analyses are available in our open-source repository.

\section{Pareto Front Visualizations}
\label{sec:appendix-pareto-front}

\begin{figure*}[ht!]
\centering
\begin{subfigure}{0.49\linewidth}
    \includegraphics[width=\linewidth]{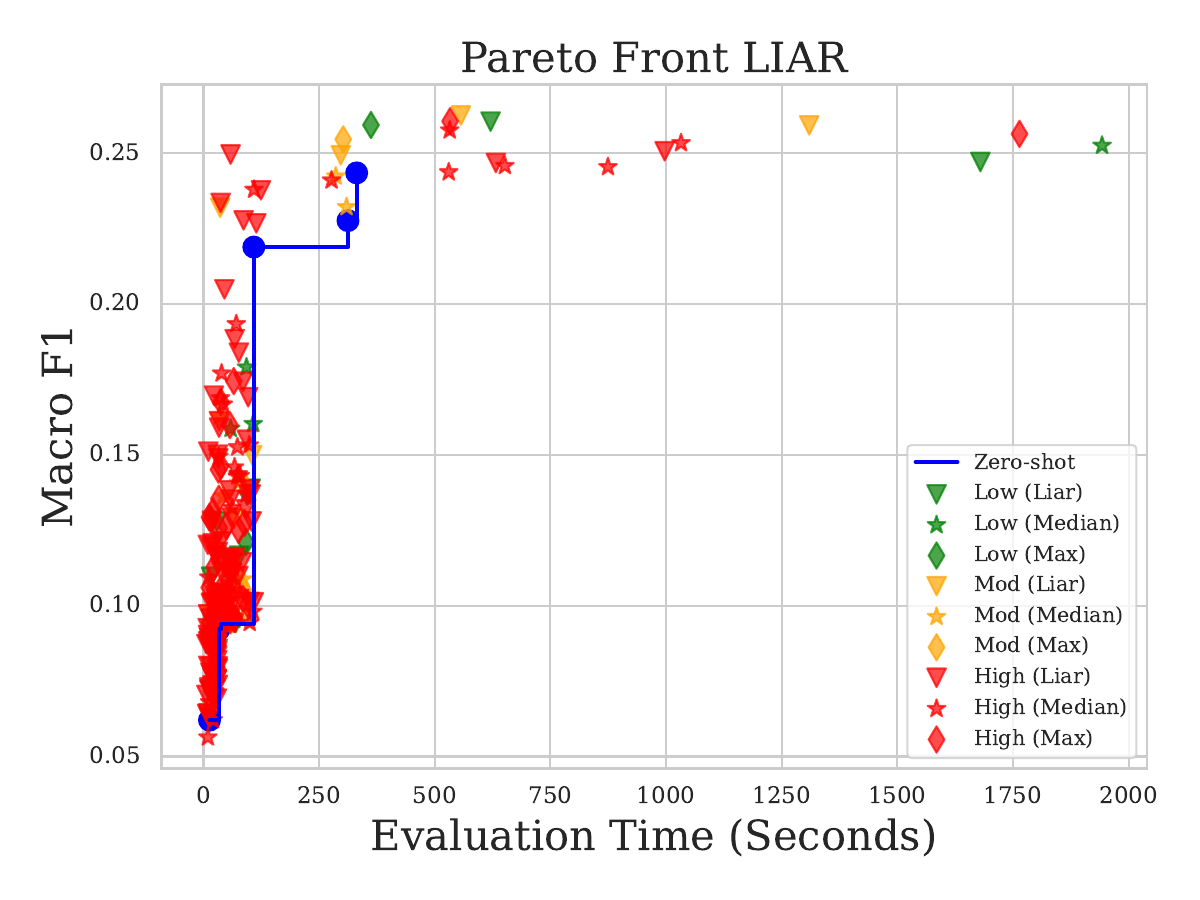}
    \caption{}
    \label{fig:pareto-liar}
\end{subfigure}
   \hfill
\begin{subfigure}{0.49\linewidth}
  \includegraphics[width=\linewidth]{images/pareto_fronts/pareto_front_sst2.pdf}
    \caption{}
    \label{fig:pareto-sst2}
\end{subfigure}
   \hfill
\begin{subfigure}{0.49\linewidth}
    \includegraphics[width=\linewidth]{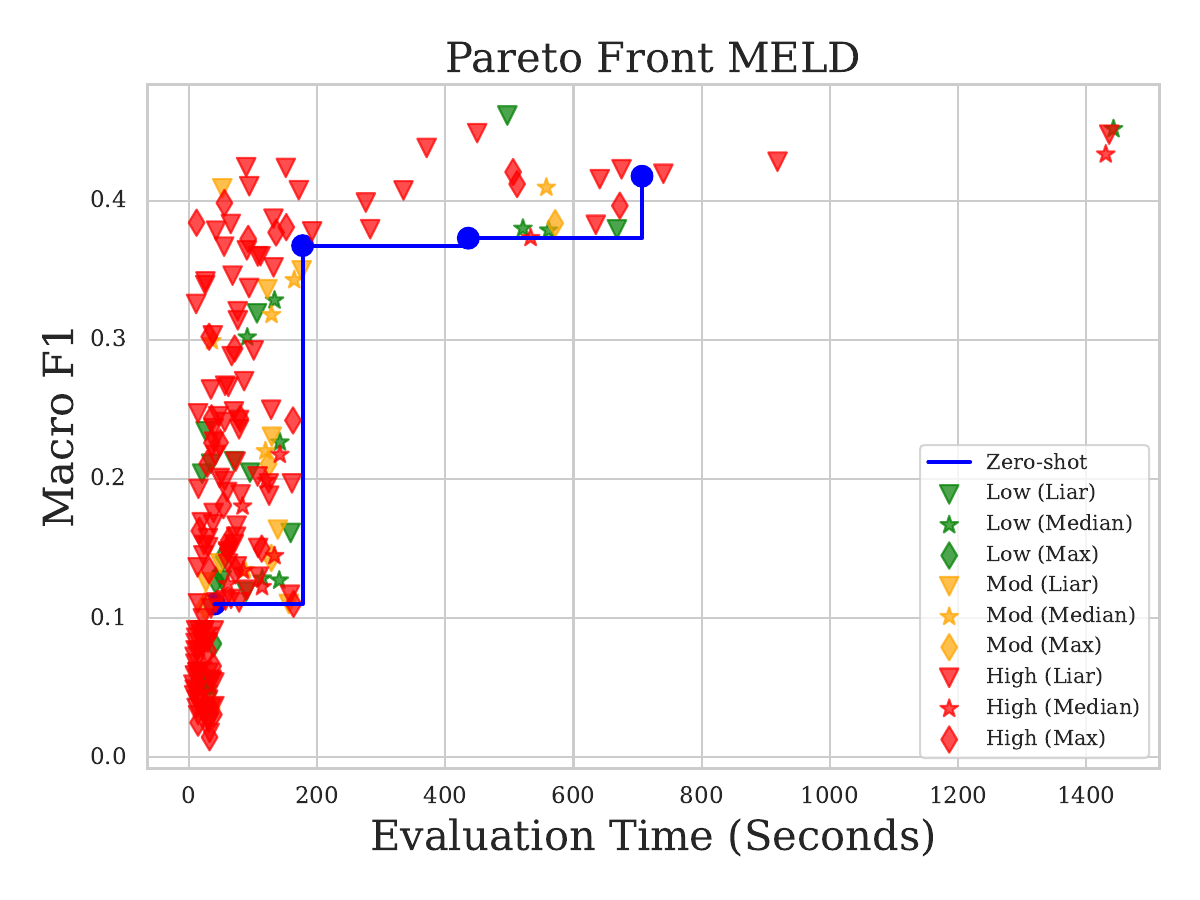}
    \caption{}
    \label{fig:pareto-meld}
\end{subfigure}
   \hfill
\begin{subfigure}{0.49\linewidth}
  \includegraphics[width=\linewidth]{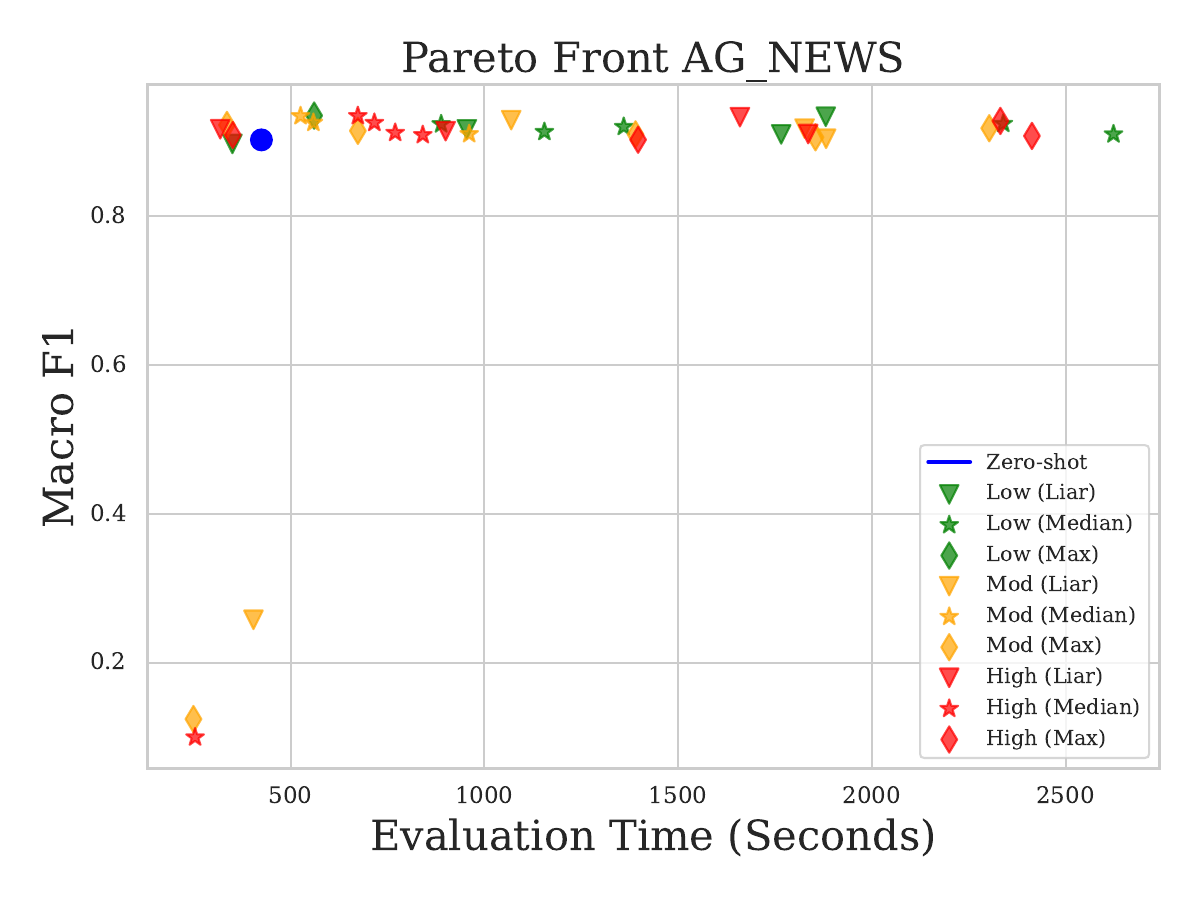}
    \caption{}
    \label{fig:pareto-ag-news}
\end{subfigure}
   \hfill
\begin{subfigure}{0.49\linewidth}
  \includegraphics[width=\linewidth]{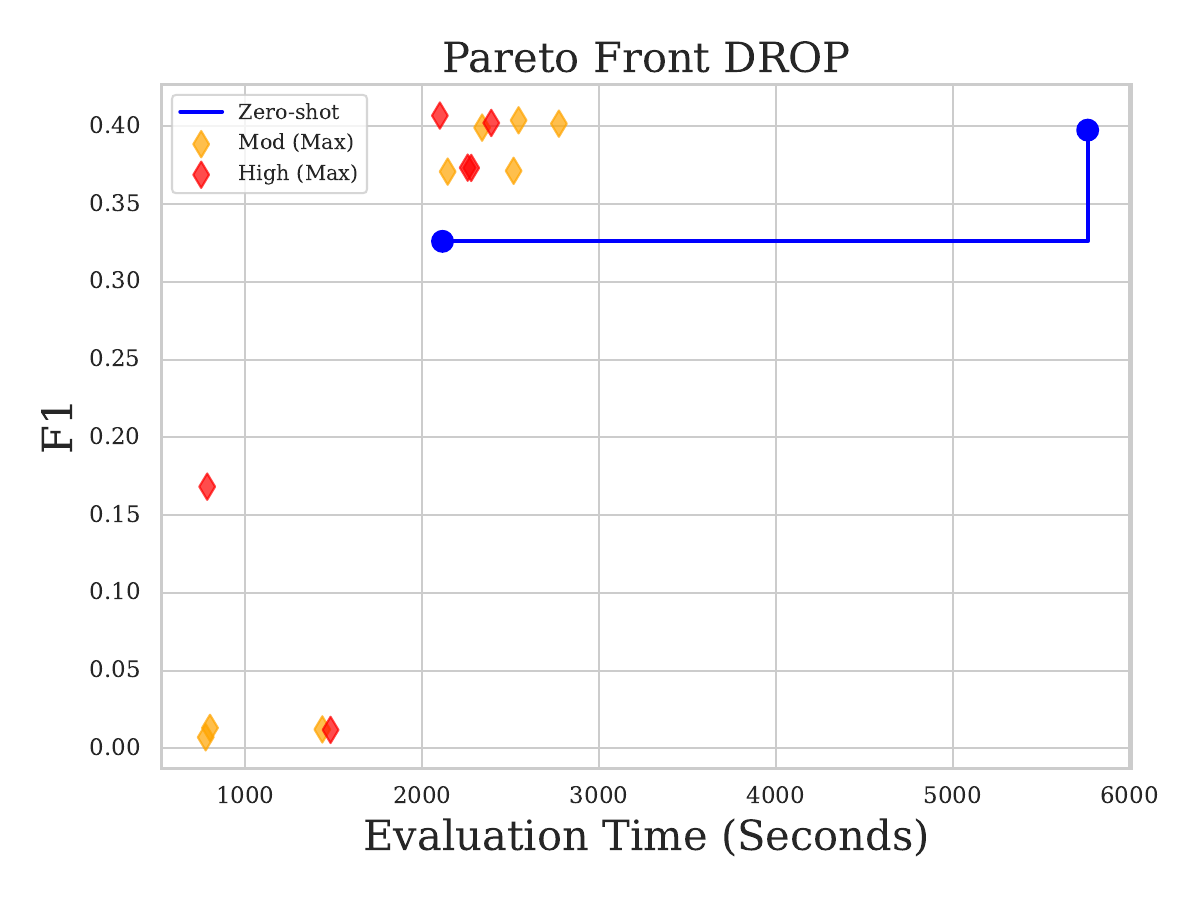}
    \caption{}
    \label{fig:pareto-drop}
\end{subfigure}
   \hfill
\begin{subfigure}{0.49\linewidth}
  \includegraphics[width=\linewidth]{images/pareto_fronts/pareto_front_squad.pdf}
    \caption{}
    \label{fig:pareto-squad}
\end{subfigure}

  \caption {Comparison of Pareto fronts for zero-shot baseline (solid blue line) and warm-start priors at Low (green), Moderate (orange), and High (red) bias levels. Each point plots $(ET, F1_{\text{macro}})$ for classification tasks (a–d) or $(ET, F1)$ for QA tasks (e–f). Points to the left or above the baseline outperforms the zero-shot Pareto front.}
  \label{fig:pareto}
\end{figure*}

Figure~\ref{fig:pareto} presents the Pareto fronts obtained on each benchmark under the zero-shot baseline and three representative warm-start bias levels (Low, Moderate, High).

Across all datasets, warm-start priors shift the search toward regions that often dominate zero-shot pipelines in both evaluation time ($ET$) and task performance ($F1_{\text{macro}}$ or $F1$). Below we highlight key observations:
Points that lie \emph{to the left of} or \emph{above} the baseline front dominate the baseline in at least one objective. In most cases, WS solutions (e.g., \emph{High WS - Median}, \emph{Mod WS - LIAR}) simultaneously improve upon the baseline's $ET$ and $F1_{\text{macro}}$, indicating superior pipelines. Below, we discuss notable observations by dataset.

\paragraph{LIAR.} High-bias priors calibrated on LIAR produce up to 40\% of pipelines that dominate the baseline, reducing error rates by roughly sevenfold (cf. Table~\ref{tab:multi-objective}). Due to the substantial meta-feature similarity between LIAR and MELD (Figure~\ref{fig:distance-heatmaps}), both tasks see rapid convergence to high-$F1_{\text{macro}}$ regions.

\paragraph{SST2.} With fewer closely related experiences, Moderate bias yields the best trade-offs, uncovering pipelines that match or slightly exceed baseline $F1_{\text{macro}}$ in less time, demonstrating robustness against negative transfer.

\paragraph{MELD.}
Figure~\ref{fig:pareto-meld} demonstrates how \textsc{meld}, like \textsc{liar}, sees \emph{numerous} WS-discovered solutions outclassing the baseline. These configurations often exploit shared meta-features between \textsc{meld} and \textsc{liar} (see Figure~\ref{fig:distance-heatmaps}), culminating in faster convergence and higher accuracy, with fewer errors during the search. Mirroring \textsc{liar}, \textsc{High WS - LIAR} dominates, diminishing the error ratio by sevenfold and almost getting 50\% winning ratio (Figure~\ref{fig:wins-by-configs}).

\paragraph{AG News.}
Figure~\ref{fig:pareto-ag-news} shows that while \textsc{ag news} has only moderate overlap with other tasks, WS still yields solutions that meet or beat baseline performance in time-accuracy trade-offs. Notably, \textsc{mod} and \textsc{high}-bias configurations reduce error rates (see Table~\ref{tab:multi-objective} in the main text), suggesting that historical knowledge, even if partially relevant, helps prune more obviously unproductive hyperparameter regions.

\paragraph{DROP and SQuAD}
For QA, High bias priors achieve dramatic gains on SQuAD, raising $F1$ from 0.34 to 0.89 and cutting mean $ET$ by 3×. On DROP, Moderate and High priors both improve $F1$ and reduce evaluation time, confirming cross-family transfer efficacy (Table~\ref{tab:qa}).

\end{document}